%% file: main.tex
\documentclass{article}

\usepackage{microtype}
\usepackage{graphicx}
\usepackage{subcaption}
\usepackage{booktabs} 
\usepackage{hyperref}
\usepackage{tikz}
\usetikzlibrary{arrows.meta, positioning, calc}

\usepackage{multirow}

\usepackage[preprint]{icml2026}

\usepackage{amsmath, color}
\usepackage{amssymb}
\usepackage{mathtools}
\usepackage{amsthm}

\usepackage[capitalize,noabbrev]{cleveref}

\theoremstyle{plain}

\theoremstyle{definition}

\theoremstyle{remark}

\usepackage[textsize=tiny]{todonotes}

\newcommand{\tool}{\textsc{EpiAgent}}

\begin{document}

\twocolumn[
  \icmltitle{Agentic Framework for Epidemiological Modeling}

  \icmlsetsymbol{equal}{*}

  \begin{icmlauthorlist}
    \icmlauthor{Rituparna Datta}{equal,UVA}
    \icmlauthor{Zihan Guan}{equal,UVA}
    \icmlauthor{Baltazar Espinoza}{BI}
    \icmlauthor{Yiqi Su}{VT}
    \icmlauthor{Priya Pitre}{VT}
    \icmlauthor{Srini Venkatramanan}{BI}
    \icmlauthor{Naren Ramakrishnan}{VT}
    \icmlauthor{Anil Vullikanti}{UVA,BI}
  \end{icmlauthorlist}

  \icmlaffiliation{UVA}{Department of Computer Science, University of Virginia, Charlottesville, USA}
  \icmlaffiliation{VT}{Department of Computer Science, Virginia Tech, Blacksburg, USA}
  \icmlaffiliation{BI}{Biocomplexity Institute, Charlottesville, USA}

  \icmlcorrespondingauthor{Anil Vullikanti}{vsakumar@virginia.edu}


  \vskip 0.3in
]
\printAffiliationsAndNotice{} 

\begin{abstract}
Epidemic modeling is essential for public health planning, yet traditional approaches rely on fixed model classes that require manual redesign as pathogens, policies, and scenario assumptions evolve. We introduce \tool{}, an agentic framework that automatically synthesizes, calibrates, verifies, and refines epidemiological simulators by modeling disease progression as an iterative program synthesis problem. 
A central design choice is an explicit epidemiological flow graph intermediate representation that links scenario specifications to model structure and enables strong, modular correctness checks 
before code is generated. Verified flow
graphs are then compiled into mechanistic
models supporting interpretable parameter learning under physical and epidemiological
constraints.
Evaluation on epidemiological scenario case studies demonstrates that \tool{} captures complex growth dynamics and produces epidemiologically consistent counterfactual projections across varying vaccination and immune escape assumptions. Our results show that the agentic feedback loop prevents degeneration and significantly accelerates convergence toward valid models by mimicking professional expert workflows.\end{abstract}

\input{intro}

\input{relatedwork}
\input{problem_statement}
\input{method}

\input{implementation}
\input{ablation}

\input{result}

\input{baseline_exp}
\input{conclusion}

  

\nocite{langley00}
\input{impactStatement}
\bibliographystyle{icml2026}
\bibliography{mainicml}

\appendix
\input{appendix}
\input{agent_table}
\input{agent_result_table}
\input{discussion}

\input{limitations}
\end{document}

%% file: intro.tex
\section{Introduction}
\begin{figure*}[htbp]
    \centering
    \includegraphics[width=0.95\linewidth]{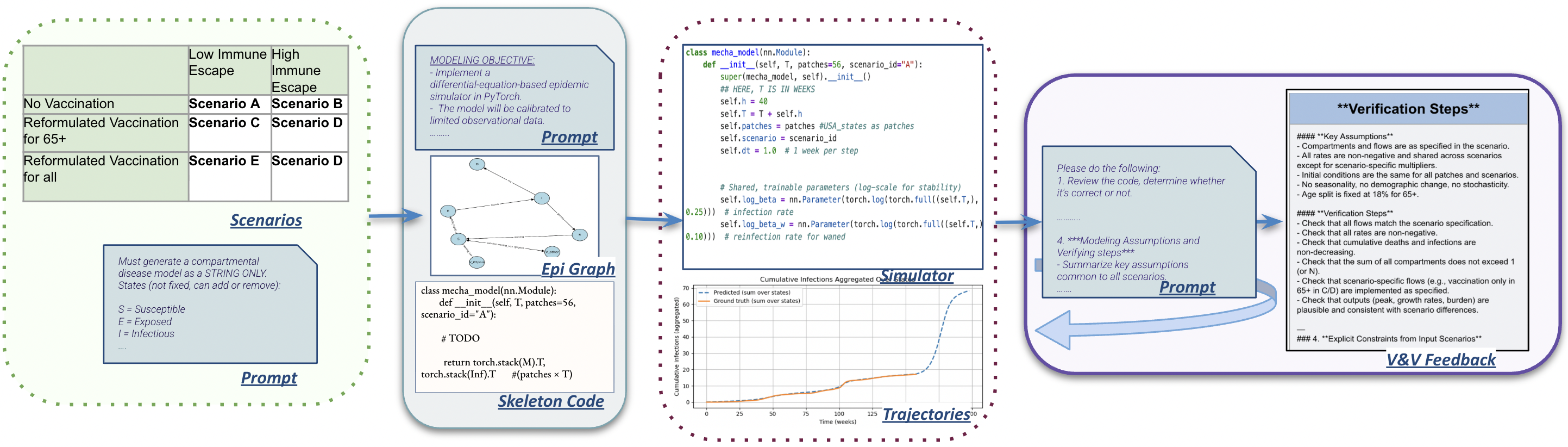}
\caption{
Data flow in \tool{} has the following structure:
\textit{Prompt} $\;\leftrightarrow\;$ \textit{Flow Graph}
$\;\rightarrow\;$ \textit{Simulator Code}
$\;\rightarrow\;$ \textit{Results}. 
The flow graph is an abstract representation of the epidemic model being learned, and supports easier verification.
The flow graph is then transformed into the actual epidemic model and is calibrated, and fully verified.
The components here map to those in the architecture of \tool{}, shown in Figure \ref{fig:pipeline}.
}
\vspace{-0.5em}
    \label{fig:workflow}
\end{figure*}
Epidemic modeling is a core tool for public health planning, providing the analytical framework necessary for policymakers to navigate disease dynamics and evaluate intervention strategies, e.g., \cite{hethcote2000mathematics,brauer2012mathematical,anderson1991infectious, marathe:cacm13,espinoza2023coupled}. 
During the COVID-19 pandemic, these models proved indispensable for guiding high-stakes decisions and informing public health policy~\cite{loo2024us}. 
A typical epidemiological model encodes assumptions about disease transmission, immunity, and population stratification and structure.
Simulation-based analyses of these models, which are calibrated using available data (usually limited and noisy), are used to study projections and provide insights about different policy scenarios.
The COVID-19 Scenario Modeling Hub (SMH) demonstrates the value of multi-model ensembles in providing robust, long-term projections to guide federal and state-level responses~\cite{midas-covid19-scenario-modeling-hub, reich2022collaborative}. Since such projections directly inform resource allocation and clinical policy, the integrity, interpretability, and reliability of the modeling process are paramount. Such efforts have expanded since to support seasonal projections across multiple pathogens, including influenza and RSV. Multiple modeling frameworks such as agent-based models~\cite{chen2024role} and metapopulation models~\cite{porebski2024data} have been used to support this effort, ranging across scenarios pertaining to pathogen characteristics~\cite{truelove2022projected}, vaccination recommendations~\cite{jung2024potential,loo2025scenario}, and non-pharmaceutical interventions~\cite{borchering2021modeling}.

Epidemiological modeling pipelines today rely heavily on manual model design and expert intervention. 
When policies change, new variants emerge, or immunity assumptions shift, epidemiologists must redesign model structures, revise transition logic, and re-implement simulators---an iterative and time-consuming process that limits scalability and responsiveness.
Large Language Models (LLMs) and agentic AI systems have demonstrated significant potential in automating complex scientific and engineering workflows through iterative generation and refinement, e.g., novel symbolic regression strategies to learn complex models from data~\cite{grayeli2024symbolic}. 
Agent-based frameworks have already been  deployed in fields such as code synthesis \cite{zhang2024pair,wang2024planning,hong2023metagpt} and automated research laboratories \cite{schmidgall2025agent}. 
In these systems, structured artifacts—--such as simulation codes or experimental protocols--—are generated, and corrected through autonomous feedback loops \cite{lu2024ai}. 
However, such methods have not been explored for epidemiological models, a significantly more sophisticated task than the aforementioned
science applications.
The motivating question for our work is thus: \emph{can agentic systems be developed to assist in epidemiological modeling and analysis?}

We show how naively prompting LLMs to generate compartmental models and simulators leads to frequent structural inconsistencies and fragile behavior.
We introduce \tool{}, an agentic framework for constructing epidemic models from natural-language epidemic study descriptions;
a crucial component in \tool{} is a flow graph representation, which helps improve the robustness of the learned models, as illustrated in Figure~\ref{fig:workflow}.
Our main contributions are: \textbf{(1.)} We introduce \emph{retrieval-augmented flow graph synthesis}, which bridges scenario specifications and mathematical structure by generating epidemiological flow graphs that satisfy hard structural constraints, and valid compartmental transitions, and can be verified more easily. \textbf{(2.)} We present a \emph{graph-to-model compilation} mechanism that transforms verified flow graphs into mechanistic simulators 
providing an explicit structural scaffold for disease parameter learning with the real-world data. \textbf{(3.)} We design a \emph{multi-agent verification and validation architecture} in which specialized agents enforce correctness properties—mathematical validity, scenario fidelity, and mechanistic interpretability—and provide actionable feedback, mirroring expert epidemiological modeling workflows.
\textbf{(4.)} We evaluate \tool{} on epidemiology modeling studies \cite{midas-covid19-scenario-modeling-hub}, and a behavioral SEIR baseline \cite{gozzi2025comparative}, demonstrating that the generated simulators achieve accurate fits to observed trajectories, produce epidemiologically consistent counterfactual projections, and converge to valid model structures more reliably than unguided generation.

%% file: relatedwork.tex
\section{Related Work}

\paragraph{Epidemiological Modeling.}
This is a core tool for public health planning \cite{
marathe:cacm13,venkatramanan2019optimizing,espinoza2020mobility,mahmud2025adaptive,dixit2023airborne}, since data is usually limited, and policy analysis questions cannot be solved simply using machine learning methods.
Large-scale efforts such as the COVID-19 Scenario Modeling Hub demonstrate the value of ensemble-based evaluation across diverse mechanistic and statistical models \cite{midas-covid19-scenario-modeling-hub,howerton2023evaluation,shea2023multiple}. These efforts also highlight the challenge of maintaining structural consistency and interpretability under distribution shift, as evolving policies, variants, and immunity assumptions often require manual redesign of model structure rather than simple parameter re-estimation.

\paragraph{Discovering equations and physical models from data.}
This class of work includes symbolic regression, sparse identification, and system identification aims to recover governing equations from data \cite{brunton2016discovering,rudy2017data,champion2019data,ljung1987theory,brunton2021modern, shojaee2023transformer}. 
Another related topic involves neural ODEs and physics-informed neural networks, which learn continuous-time dynamics with neural components \cite{chen2018neural,raissi2019physics,karniadakis2021physics,rackauckas2020universal,cuomo2022scientific}. However, epidemic simulators differ in important ways: they encode structured compartment transitions and scenario-driven intervention logic, and are often used for counterfactual and mechanistic reasoning, where black-box components and fixed topologies can hinder interpretability and scientific validity, even when trajectory fits are accurate.
\paragraph{Agentic Scientific Modeling Systems.}
Recent work has explored agentic and automated scientific modeling systems that leverage LLMs for program synthesis, repair, and iterative refinement through execution feedback and tool use~\cite{yao2022react,madaan2023self,wang2024planning,zhang2024pair,guan2026nimmgenlearningneuralintegratedmechanistic}, as well as broader 'AI scientist' frameworks for automating research workflows \cite{lu2024ai,schmidgall2025agent} and symbolic regression \cite{grayeli2024symbolic}. 
These systems primarily target syntactic correctness and empirical performance; typically lack domain-specific mechanistic constraints or support structural revision as a first-class operation.
\tool{} addresses this gap by treating epidemic simulator construction as an agentic, scenario-conditioned process that explicitly represents and verifies mechanistic structure, integrating verification and validation (V\&V) into the generation loop to support reliable counterfactual analysis.

%% file: problem_statement.tex
\section{Problem Statement}
\begin{figure*}[htbp]
   \centering
\includegraphics[width=0.95\linewidth, height=6cm]{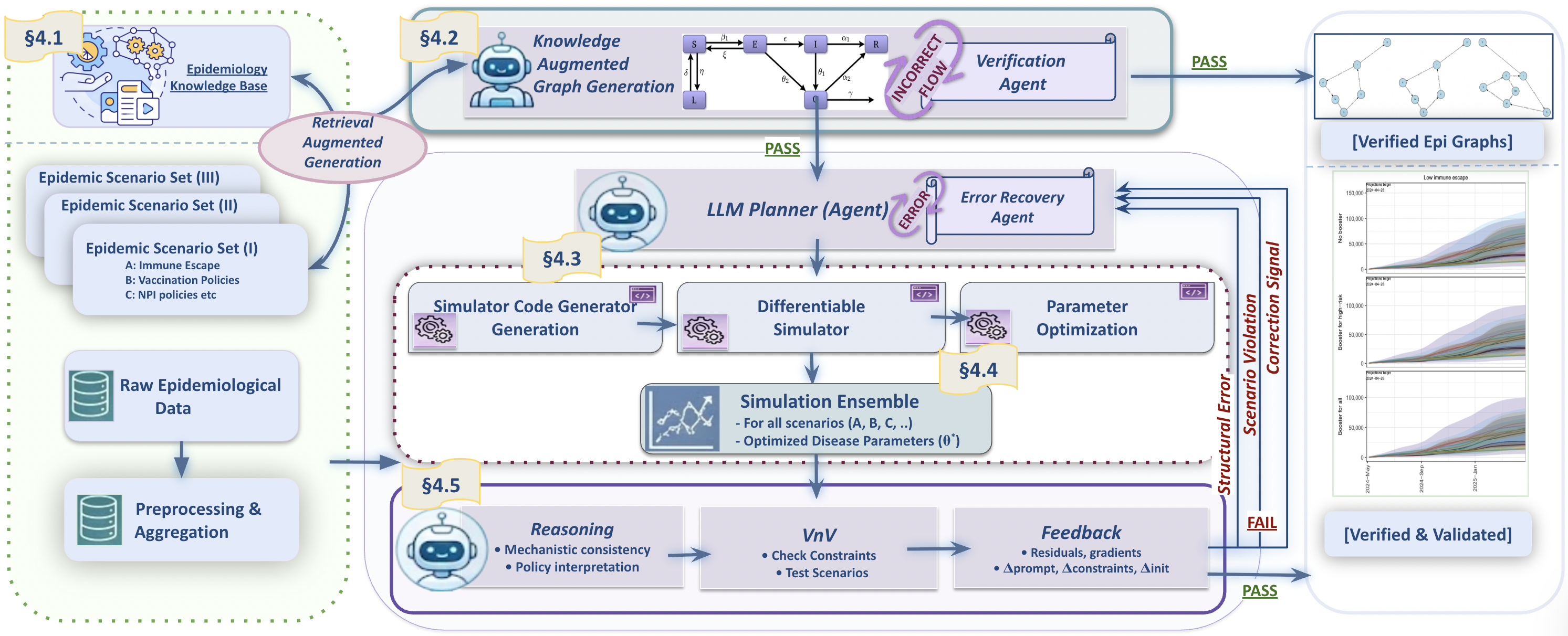}
       \caption{Agentic pipeline for epidemic scenario modeling and ensemble simulation. Natural-language scenarios are augmented with domain knowledge (\S\ref{ssec:method-scenario}), to produce prompts that guide flow-graph synthesis (\S\ref{ssec:method-graph}). Generated graphs are iteratively verified to enforce valid compartmental structure and transitions. Given a verified graph and scenario description, an LLM planner instantiates executable simulator code with automated error recovery (\S\ref{ssec:method-simulator}). Simulators are calibrated on observed data and evaluated as a scenario ensemble (\S\ref{ssec:method_train}). A multi-agent verification and validation stage enforces epidemiological and scenario-consistency constraints, retaining only structurally and behaviorally valid models (\S\ref{ssec:method_vnv}).}
       \vspace{-0.5em}
       \label{fig:pipeline}
\end{figure*}
\subsection{Preliminaries}

A large number of models are used in epidemiology, here, we focus on compartmental epidemic models, \cite{
hethcote2000mathematics,brauer2012mathematical,anderson1991infectious}.
A canonical example is the SEIR model, which partitions the population into susceptible $S(t)$, exposed $E(t)$, infectious $I(t)$, and recovered $R(t)$:
\[
\tfrac{dS}{dt}=-\beta\tfrac{SI}{N},\;
\tfrac{dE}{dt}=\beta\tfrac{SI}{N}-\sigma E,\;
\tfrac{dI}{dt}=\sigma E-\gamma I,\;
\tfrac{dR}{dt}=\gamma I
\]
where $N = S + E + I + R$ is the total population, $\beta$ is the transmission rate, $\sigma$ is the incubation rate, and $\gamma$ is the recovery rate. 
Most epidemic analyses involve  introducing additional states and transitions 
(e.g., to represent interventions, immunity, and population heterogeneity).
We refer to a specific epidemiological study as a \emph{scenario}, denoted by $s$.
This could include  (i) assumptions about disease-dynamic (e.g.,
variants, hospitalization), (ii) interventions
(e.g., timing, intensity, and compliance of non-pharmaceutical policies),
and (iii) population heterogeneity (e.g., spatial structure, demographic variation, or
contact patterns).
Let $\mathcal{X}$ denote a partitioning of the population into sub-populations (e.g., based on age groups), referred to as patches.
Let $V$ denote the set of health states being represented over all patches in $\mathcal{X}$, and let $\mathbf{x}(t)\in\mathbb{R}^K$ denote the vector of epidemic state values (e.g., fraction of susceptible and infectious people), where $K=|V|$.
An epidemic model for scenario $s$, denoted by $f_\theta^{s}(\mathbf{x}(t))$, provides the state vector values at time $t+1$. $\theta$ encodes disease and intervention parameters–may include parameters such as $(\beta, \sigma, \gamma)$, or time-varying parameters such as $\beta(t)$ or $\gamma(t)$ to capture behavioral or seasonal effects, or structured parameter functions driven by exogenous covariates. 
Some studies, such as the CDC Scenario Modeling Hub \cite{midas-covid19-scenario-modeling-hub}, require developing models for a \emph{set of scenarios}, denoted $\mathcal{S}$.
There is significant theory that can be leveraged about validity of mechanistic models, e.g., non-negativity of parameters, and  monotonicity of cumulative quantities, e.g., \cite{brauer2012mathematical}.
A simulation for a model $f^s_{\theta}(\cdot)$ refers to the actual implementation of the dynamical system model.

To better understand model fit and expressive capacity, we also consider \emph{hybrid composition models}. In these models, the parameter set $\theta$ is learned using a neural network \cite{chopra2023differentiable,guan2025framework,datta2025calypso}.
Such hybrid models increase flexibility when fitting complex or multi-peak epidemic trajectories, but are hard to interpret due to the black box component. 
Here, we include such models and study the improvement in fit by neural augmentation, but this is not the main focus of our work.

\noindent
\textbf{Scenario Flow Graph.}
Each scenario $s \in \mathcal{S}$ can be represented by a directed epidemiological flow graph, $G = (V, E)$, on the nodes in $V$, where each edge $(u,v) \in E$ denotes an admissible transition, and is parameterized by shared disease parameters $\theta$. 
The SEIR model incorporating disease-induced mortality can be represented by the directed flow graph

$
S \xrightarrow{\;\beta\;} E
\;\xrightarrow{\;\sigma\;}
I
\;\xrightarrow{\;\gamma\;}
R,
\qquad
I
\;\xrightarrow{\;\mu\;}
D
$

where $S$, $E$, $I$, $R$, and $D$ denote the susceptible, exposed, infectious, recovered, and deceased compartments, respectively. The parameters $\beta$, $\sigma$, $\gamma$, and $\mu$ correspond to the transmission, progression, recovery, and disease-induced death rates~\cite{brauer2012mathematical}.

\subsection{Problem Definition}
Let
$\mathcal{D}
=
\big\{
x_i^{(s)}(p,t), i\in V'
\big\}_{s=1,t=1}^{\mathcal{S},T}$
denote the dataset of observed epidemiological time series data for a subset of node states $V'$ for each  patch $p \in \mathcal{X}$ over discrete weekly time steps
$t = 1,\ldots,T$.
We restrict the model parameters and the compartmental flows to a feasible set of constraints
$\mathcal{C} \subseteq \Theta$, defined as 
\vspace{-0.5em}
\begin{equation}
\label{eq:constraints}
\mathcal{C}
=
\Big\{
\theta \in \Theta
\;\big|\;
\mathbf{g}(\theta) \le \mathbf{0},\;
\mathbf{h}(\theta) = \mathbf{0}
\Big\},
\end{equation}
where $\mathbf{g}(\theta)$ encodes inequality constraints such as disease parameter
positivity, boundedness, and numerical stability conditions, and $\mathbf{h}(\theta)$ encodes
equality constraints such as conservation laws and normalization constraints---incorporated through
$\mathcal{C}$.
Given $\mathcal{D}$ and $\mathcal{S}$, learning an epidemic model corresponds to
solving the following optimization problem:
\vspace{-1em}
\begin{equation}
\small
\label{eq:loss_func}
\theta^{\star}
=
\arg\min_{\theta \in \mathcal{C}}
\sum_{p=1}^{P}
\sum_{t=1}^{T}
\mathcal{L}\!\left(
f_{\theta}^{\mathcal{S}}\!(\cdot),
\;
\big(
x_i^{(s)}(p,t)
\big)
\right) \vspace{-2em}
\end{equation}

where $\mathcal{L}(\cdot)$ denotes loss function that measures the discrepancy between
simulated and observed trajectories.

Given a set of scenarios $\mathcal{S}$, specified using natural language description and observational data $\mathcal{D}$, the goal of \tool{} is to produce epidemiological models $f^s_{\theta}, s\in\mathcal{S}$, and the associated simulations, satisfying the following properties:
(1) \emph{Satisfactory accuracy}: the generated models should minimize the empirical loss as formalized by the optimization objective in Eq~\eqref{eq:loss_func}.
(2) \emph{Model validation and verification}: the learned models $f^s_{\theta}, s\in\mathcal{S}$ should satisfy constraints from epidemiology theory (Eq~\ref{eq:constraints}, details in \S~\ref{ssec:method_vnv}).
and
(3) \emph{Scenario consistency and validity}: the learned models should be internally consistent across scenarios and faithfully reflect the assumptions, interventions, and constraints specified by each scenario description.



%% file: method.tex
\section{Method}
We introduce \tool{}, an agentic framework for epidemiological modeling. 
Figure~\ref{fig:workflow} presents a high-level view of the system’s data flow from natural-language scenario prompts to simulation outputs, while Figure~\ref{fig:pipeline} details the internal agentic pipeline that implements this architecture. 
Below, we describe the main components of \tool{}.



\subsection{Scenario Parsing and Knowledge Retrieval}\label{ssec:method-scenario}

The first phase of \tool{} takes a natural-language scenario specification $\mathcal{S}$ as input and outputs a knowledge-augmented user-generated prompt encoding explicit structural and scenario constraints $\mathcal{C}$ for flow-graph generation.


For extracting relevant knowledge, epidemic descriptions are embedded using a sentence-transformer model and augmented with relevant epidemiological knowledge retrieved from a curated corpus using a FAISS-based vector index \cite{reimers-2019-sentence-bert,douze2024faiss}. Retrieved passages are combined with the specification to form a knowledge-augmented prompt, grounding model construction in established epidemiological theory \cite{gao2023retrieval}.


\subsection{Knowledge-Augmented Synthesis of Epidemiological Flow Graphs}\label{ssec:method-graph}
The construction of the epidemiological flow graph $\mathcal{G} = (V, E)$ serves as the structural foundation for the simulator. 
We frame flow-graph generation as an iterative, knowledge-conditioned synthesis process guided by epidemiological constraints and scenario assumptions.
The synthesized graph undergoes an iterative verification loop to ensure "correctness" through hard structural constraints on admissible compartments and transitions.

\noindent
Graphs that violate these constraints are rejected and regenerated using explicit error feedback. Graphs that pass these checks are then subjected to an \emph{agentic verification} loop that evaluates scenario consistency and requires epidemiological justification of the proposed transitions. During this step, structured feedback is injected to correct specific deficiencies (e.g., missing waning transitions, invalid vaccination flows, or inconsistent compartment semantics), producing a revised graph. Verification is automated by (i) checking alignment between compartments and scenario assumptions (e.g., vaccination eligibility, immune escape, and waning), and (ii) enforcing explicit epidemiological justification for each transition. The loop terminates only when the graph is verified as valid under both structural and scenario-specific criteria. If verified as valid, the loop terminates. This graph-level verification prevents invalid structures from propagating into executable code. A sample graph-verification response is shown in Appendix Figure~\ref{fig:graphVerfication}.


\subsection{Dynamical System Model and Simulator Generation} \label{ssec:method-simulator}

In this step, given a verified flow graph and disease specifications, an agentic system instantiates dynamical models and its executable simulator, $f_{\theta}^{s}(\cdot)$.
The form of the resulting system of ordinary differential equations is determined by the verified graph. Executable simulator code is then automatically generated and calibrated against ground-truth.


\noindent
\textbf{Skeleton Code and Execution Interface.} To ensure executability and reproducibility, we provide the model with a code skeleton, $\mathcal{CK}$, that fixes the execution interface while leaving epidemiological logic unconstrained. 
The skeleton defines environment-dependent components—such as module structure, time indexing, and scenario identifiers—and requires the simulator to implement a differentiable forward pass that outputs scenario-specific target trajectories (e.g., infections and death counts) (Fig \ref{fig:skeletonCode}, Appendix). 
By separating the fixed execution interface from the flexible model logic, we enable automated execution, training, and validation without manual intervention.

\noindent
\textbf{Constraint-Guided Prompting.} Alongside the skeleton code, we impose a set of hard constraints $\mathcal{C}$ that guide generation and prevent common failure modes. These constraints can either be derived from external epidemiological knowledge or hardcoded, and fall into two categories: (i) constraints enforcing valid epidemic structure, such as numerical stability and compartmental consistency, and (ii) execution constraints that prevent unsafe coding behaviors, including reassignment of trainable parameters within the forward pass. We deliberately avoid enforcing parameter non-negativity through ad hoc clipping or rectification; instead, violations are treated as signals that the model structure itself must be revised. Epidemiological principles are additionally provided as soft constraints to encourage interpretable designs without limiting expressiveness. Together, these constraints ensure that generated simulators are both executable and scientifically meaningful.

\noindent
\textbf{LLM Planner Agent, $\mathcal{A}_{plan}$.} This planning agent generates differentiable simulator code by instantiating predefined execution skeletons with compartmental structures derived from the verified flow graph, conditioned on the given specifications.
It has an \emph{Error Recovery Module} which automatically monitors the execution of synthesized code during compilation, simulation, and training. When runtime, numerical, or structural errors occur (e.g., shape mismatches, invalid tensor operations, solver instability, or violation of differentiability constraints), the agent captures the full error trace and returns it as structured feedback to the coding agent. The coding agent is then prompted to revise the implementation while preserving the prescribed model skeleton and scenario constraints. This recovery loop is executed for a bounded number of retries, ensuring robustness of autonomous code generation without manual intervention.

\subsection{Training and Calibration}
\label{ssec:method_train}


For each generated simulator, disease parameters $\theta$ are calibrated to observational data using gradient-based optimization. We use a mean squared error (MSE) loss over infection and deceased levels and first-order differences:

$
\mathcal{L}
=
\frac{1}{P}
\sum_{p=1}^P
\Big(
\mathrm{MSE}(\hat{x}_p, x_p)
+
\mathrm{MSE}(\Delta \hat{x}_p, \Delta x_p)
\Big)$

Where, $\Delta x_t = x_{t+1}-x_t$. 
Parameters are shared across scenarios unless explicitly overridden by the scenario specification, ensuring that scenario-dependent differences arise from modeled mechanisms rather than independent re-fitting. The calibrated simulator is then evaluated by generating projections across all
scenarios.
Optimization hyperparameters, including the optimizer and learning-rate schedule, are fixed and predefined; the agent is restricted to completing the simulator skeleton and cannot modify the training environment.

\subsection{Multi-Agent Verification and Validation (V\&V)}\label{ssec:method_vnv}
V\&V is enforced through automated LLM-driven modules.

\noindent
\textbf{Verification Module.} This module checks the structural and numerical correctness of the simulator, including:
(i) consistency between the executable code $f^s_{\theta}(\cdot)$ and the verified flow graph $G$,
(ii) non-negativity of states and parameters: For all compartments $x_i(t) \ge 0$ for all $t$, preventing unphysical population counts,
(iii) conservation of population mass: The system must satisfy $\sum_{i} \frac{dx_i}{dt} = 0$, ensuring the total population remains constant, \emph{excluding deceased flows}, and
(iv) numerical stability during simulation.

\noindent
\textbf{Model Reasoning Module.} This module analyzes and explains the generated code and model structure, providing explicit justifications for compartment choices, parameterization, and dynamical assumptions, thereby improving interpretability and transparency.

\noindent
\textbf{Validation Module}. This module evaluates plausibility and scenario fidelity, assessing whether simulated outcomes reflect intended scenario mechanisms and cross-scenario differences across all evaluated scenarios.

\noindent
Finally, the \textbf{Performance Feedback Module} evaluates calibration and validation results (e.g., loss trends, generalization gaps) and produces targeted feedback to guide iterative model refinement toward improved predictive performance.

The full procedure is formalized in Algorithm~\ref{alg:agentic-epi}, Appendix.

%% file: implementation.tex
\section{Experimental Setup}
\textbf{Datasets.} We evaluate our framework using scenario modeling case studies from the COVID-19 Scenario Modeling Hub archive, which comprises 19 scenario rounds (\emph{Case Study I}). Each round specifies structured assumptions over interventions, vaccination, and immunity, together with corresponding epidemiological time series for projection and evaluation~\cite{midas-covid19-scenario-modeling-hub}. We additionally consider \emph{Case Study II}, which evaluates calibration and probabilistic projection on real-world epidemiological data using a standard age-structured behavioral SEIR model. This study uses publicly available COVID-19 surveillance data from four heterogeneous locations, including reported deaths, mobility-derived behavioral signals, and seasonality covariates~\cite{gozzi2025comparative}. Compartmental structures and transition semantics are derived from established formulations in mathematical epidemiology~\cite{brauer2012mathematical}, which serve as the epidemiological knowledge base queried via retrieval-augmented generation (RAG).

\textbf{LLM Selection.} Epidemiological flow graphs are generated using \texttt{gpt-4o-mini}\cite{achiam2023gpt}. Knowledge retrieval is implemented using sentence-transformer embeddings (\texttt{all-MiniLM-L6-v2}) indexed with FAISS, which enables efficient semantic retrieval from epidemiology textbooks. We use \texttt{gpt-4.1} for planning and simulator code generation, as these stages require higher reasoning capacity, while verification, validation, and feedback agents use the lighter-weight \texttt{gpt-4.1-mini} model to reduce computational cost without sacrificing correctness.

\noindent
\textbf{Agentic Configuration.} LLM agents operate under controlled and adaptive sampling. The temperature is initialized at zero to encourage deterministic generation and is increased only when repeated or cyclic responses are detected. Token budgets, rate limits, and a fixed maximum number of retries are enforced to bound inference cost. 

\noindent
\textbf{Execution Environment.} All simulators are calibrated by completing a fixed PyTorch execution environment and are trained in a predefined setting with a fixed optimizer (Adam) and learning-rate scheduler (ReduceLROnPlateau).


\noindent
\textbf{Hardware Configuration.} All experiments are conducted using PyTorch with GPU acceleration on NVIDIA Tesla V100-SXM2 GPUs (32GB memory) with CUDA.

%% file: ablation.tex
\subsection{Ablation Study}
\label{sec:ablation}
To isolate the contribution of \tool{}’s key design choices, we conduct targeted ablations that remove (i) flow-graph verification, (ii) access to external epidemiological knowledge, and (iii) the flow-graph intermediate representation (i.e., direct scenario-to-code generation). We evaluate each variant by inspecting the validity of the induced model structure, the frequency and severity of structural errors, and the number of generate–verify iterations required to reach a mechanistically valid simulator.

\textbf{Without Flow-Graph Verification.}
Omitting graph-level verification results in epidemiologically invalid structures that propagate into dynamical models. Across our experiments, incorrect flow graphs result in incorrect scenario projections in over 87\% of cases, underscoring the necessity of explicit graph-level verification. 

\textbf{Without External Knowledge Retrieval (No RAG).}
Without epidemiological knowledge retrieval, the LLM defaults to the simplest SEIRD-style graphs, often omitting scenario-specific mechanisms such as waning immunity, or age-targeted vaccination. Empirically, knowledge guidance reduces the average number of graph-generation iterations with feedback by $\sim20\%$. (Figure \ref{fig:promptToGraph} and \ref{fig:promptToGraphwoknowledge}, Appendix).

\textbf{Without Flow-Graph Intermediate Representation (Direct Scenario-to-Code).}
Directly generating equations from scenario text leads to frequent structural errors in code, including missing compartments, duplicated states, or invalid transitions (e.g., $R \!\to\! D$) (Figure \ref{fig:wrongCode}, Appendix). These errors are difficult to detect and recover at the code level. Introducing an explicit flow-graph intermediate representation decouples structural validation from equation synthesis, enabling targeted verification and substantially improving reliability.

%% file: result.tex
\section{Results}
We evaluate the proposed agentic framework through two complementary case studies to assess calibration accuracy, structural correctness, and counterfactual reasoning. \emph{Case Study I} evaluates whether agent-generated simulators correctly interpret and execute complex counterfactual scenarios. \emph{Case Study II} evaluates whether \tool{} can recover and deploy a standard behavioral epidemiological model for fitting and probabilistic projection across real-world data. Together, these case studies evaluate \tool{} by addressing the following questions.
\\
1. Can \tool{} calibrate epidemic dynamics? (\S\ref{ssec:results-calibration})\\
2. Does graph verification avert incorrect solutions? (\S \ref{ssec:incorrectFlow})\\
3. Does code-level verification prevent degenerate low-loss solutions (e.g., negative states, mass imbalance)? (\S\ref{ssec:model-verification})\\
4. Do scenario-conditioned models respond correctly to counterfactual interventions? (\S~\ref{ssec:sc_validation}) \\
5. Does agentic feedback improve performance? (\S\ref{ssec:results-convergence}) \\
6. Can \tool{} be applied to classical epidemiological models to generate reliable projections? (\S\ref{ssec:baseline_epipaper})\\
While \tool{} is designed for general epidemiological modeling, we also use a specific COVID-19 scenario-modeling round as a primary case study to demonstrate its counterfactual reasoning capabilities.
\begin{table}[h]
\centering
\caption{Performance comparison of modeling approaches.}
\label{tab:model_comparison}
\small
\setlength{\tabcolsep}{3.5pt}
\resizebox{0.95\linewidth}{!}{%
\begin{tabular}{l|lccc}
\toprule
 & \textbf{Model} & \textbf{MAE} & \textbf{MSE} & \textbf{RMSE} \\
\midrule
\multirow{3}{*}{\textbf{State-avg}} 
& Time-invariant dynamical 
& 3.84$\times10^{-3}$ 
& 3.10$\times10^{-5}$ 
& 5.50$\times10^{-3}$ \\
& Time-variant ($\beta_t,\gamma_t$) 
& 8.90$\times10^{-4}$ 
& 1.30$\times10^{-6}$ 
& 1.10$\times10^{-3}$ \\
& Neural-incorporated 
& 6.27$\times10^{-4}$ 
& 7.50$\times10^{-7}$ 
& 8.00$\times10^{-4}$ \\
\midrule
\multirow{3}{*}{\textbf{Nationwide}} 
& Time-invariant dynamical 
& 2.12$\times10^{-1}$ 
& 9.57$\times10^{-2}$ 
& 3.09$\times10^{-1}$ \\
& Time-variant ($\beta_t,\gamma_t$) 
& 4.26$\times10^{-2}$ 
& 2.53$\times10^{-3}$ 
& 5.03$\times10^{-2}$ \\
& Neural-incorporated 
& 2.22$\times10^{-2}$ 
& 7.13$\times10^{-4}$ 
& 2.67$\times10^{-2}$ \\
\bottomrule
\end{tabular}
}
\vspace{-1em}
\label{tab:loss_metric}
\end{table}
\subsection{Empirical Calibration of Generated Simulators} \label{ssec:results-calibration}
\begin{figure}[h]
    \centering
\includegraphics[width=0.95\linewidth]{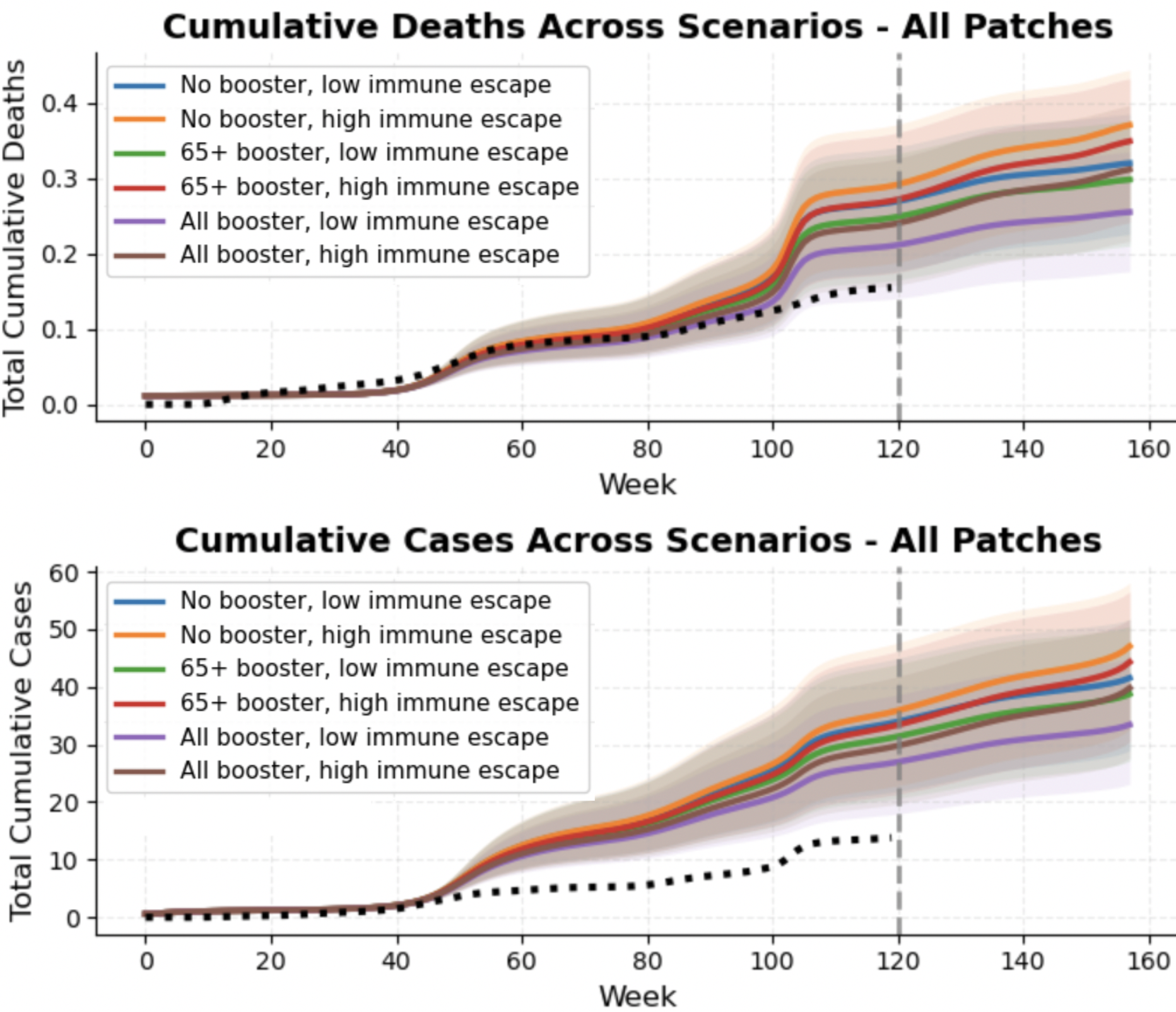}
    \caption{Cumulative COVID-19 infections aggregated at the national level across six scenarios from \cite{midas-covid19-scenario-modeling-hub}. State-level trajectories are normalized by population and aggregated to produce a single nationwide infection and death trajectory.}
    \label{fig:sc17_correct}
    \vspace{-1em}
\end{figure}

We first evaluate the empirical calibration of the simulators produced by \tool{} by comparing their projected trajectories against observed epidemiological data. 
Figure \ref{fig:sc17_correct} shows the fit of the simulator with time variant disease parameters with uncertainty.
Allowing time-varying disease parameters (e.g., $\beta(t)$, $\gamma(t)$) improves short-term forecasting accuracy 
(Table~\ref{tab:loss_metric}). 
When additional flexibility is required, the agent can introduce neural components to learn time-varying parameters. By default, however, the agent restricts the model to time-invariant parameters, avoiding unnecessary over-parameterization while preserving interpretability. \emph{Hybrid formulations} are particularly well-suited for forecasting settings, where maximizing predictive accuracy is prioritized. Despite this, the generated models are able to capture key temporal characteristics, including overall growth dynamics and peak timing, which are the primary objectives of scenario-based projection. 

\subsection{Model Verification} \label{ssec:model-verification}
We evaluate the effectiveness of the proposed verification mechanism in enforcing epidemiological and structural correctness of the synthesized simulators. Verification operates at both the \emph{structural (flow-graph)} and \emph{behavioral (code execution)} levels, and serves as a filter that eliminates epidemiologically invalid models.

\noindent
\textbf{Mechanistic validity.}
Models generated without verification frequently exhibit (i) invalid state transitions (e.g., direct $S -> R$ recovery without infection), (ii) implausible causal pathways (e.g., deaths arising from susceptible), and (iii) missing required flows (e.g., lack of waning immunity transitions under scenarios specifying immune escape). Using retrieval-augmented domain knowledge, the system identifies and rejects such models at the flow-graph level, ensuring that only graphs consistent with established epidemiological principles are admitted for simulator generation. 

\emph{To enforce mechanistic validity, \tool{} explicitly separates natural-language prompting from simulator synthesis by first generating and verifying an intermediate compartmental flow graph.}

\textbf{Impact of Graph Correctness Agent.}
\label{ssec:incorrectFlow}
In \tool{}, the objective is scenario modeling rather than point forecasting; consequently, correctness of the compartment graph is essential for causal validity and counterfactual reliability. While an incorrect graph often leads to incorrect projections, this is not the only failure mode. In some cases, parameter flexibility can compensate for structural mis-specification, producing epidemic curves that appear numerically reasonable despite encoding incorrect causal mechanisms.

\begin{figure}
    \centering
    \includegraphics[width=1\linewidth]{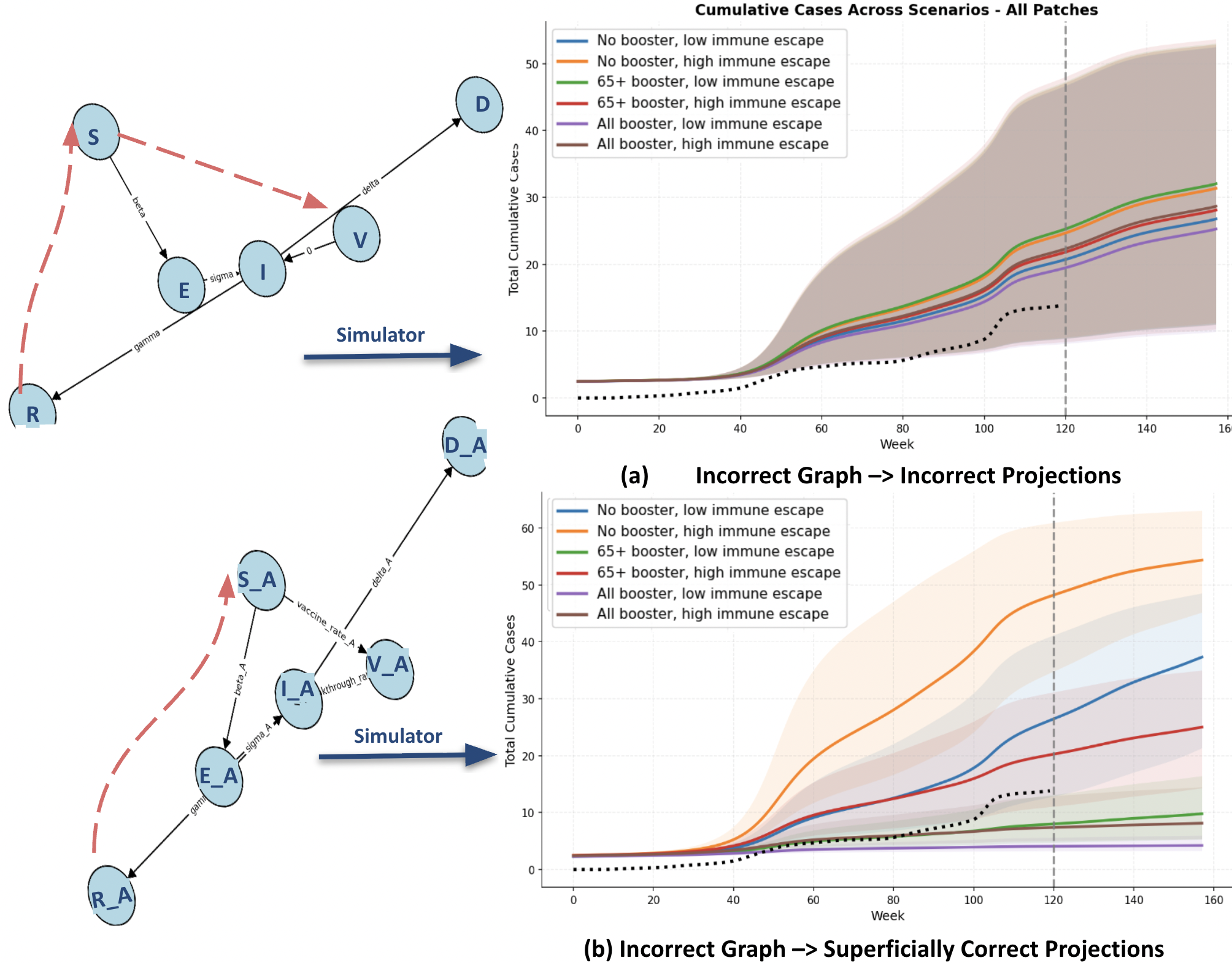}
    \caption{Error propagation from incorrect graphs to epidemic projections: may yield plausible outcomes (b) or incorrect trajectories (a)–demonstrating the necessity of structural verification.}\label{fig:graph_error_propagation}
    \vspace{-1em}
\end{figure}

We categorize graph–projection outcomes into three cases: (i) \emph{correct graph, correct result (Figure~\ref{fig:sc17_correct}}); (ii) \emph{incorrect graph, incorrect result}; and (iii) \emph{incorrect graph, correct-looking result}. Figure~\ref{fig:graph_error_propagation} illustrates the latter two cases.

In Figure~\ref{fig:graph_error_propagation}(a), the compartment graph omits population-level immunity loss (no \(R \!\to\! S\) or waning state), which structurally constrains the dynamics by suppressing reinfection. As a result, scenario projections are biased and diverge for incorrect mechanistic reasons, yielding visibly incorrect cumulative trajectories. In contrast, Figure~\ref{fig:graph_error_propagation} (b) also relies on an incorrect graph, but sometimes, flexible parameterization partially masks the structural error, yielding aggregate trajectories that appear consistent across scenarios. 

\noindent
\textbf{Parameter and state constraints.}
Verification eliminates models that violate fundamental epidemic constraints during execution. Without enforcement, gradient-based calibration can converge to numerically low-loss but epidemiologically invalid solutions. Common failure modes include: (i) negative compartment values (e.g., $I(t) < 0$ or $E(t) < 0$), which are physically meaningless; (ii) violation of population conservation, where $S(t)+E(t)+I(t)+R(t)$ exceeds or falls below the population size due to inconsistent flows; and (iii) non-monotonic cumulative quantities, such as cumulative deaths decreasing over time after calibration. The verification agent enforces non-negativity, population conservation, and monotonicity, preventing such degenerate but low-loss solutions from being accepted (Figure \ref{fig:vnvAgent}, Appendix).


\subsection{Scenario Validation} \label{ssec:sc_validation}
We validate the synthesized simulators through counterfactual analysis across scenarios. By holding the calibrated parameterization fixed and varying only the scenario specification, we examine whether projected outcomes respond in a coherent and epidemiologically interpretable manner. Differences in vaccination strategy and immune escape assumptions produce systematic shifts in projected infections, hospitalizations, and mortality, demonstrating that the models not only fit historical data but also support meaningful scenario-driven reasoning. We evaluate this capability using a structured scenario in \emph{Case Study I} (Fig \ref{fig:Covid19Sc17_description}, Appendix).

\begin{figure}[h]
    \centering
    \includegraphics[width=\linewidth]{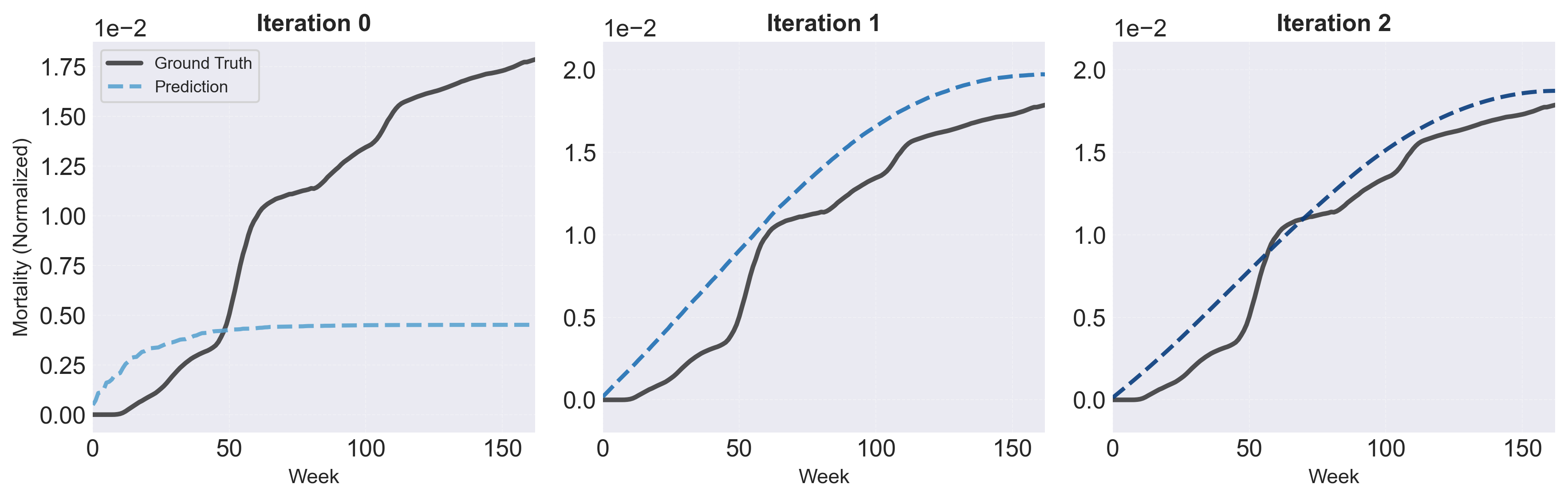}
    \caption{ Iterative refinement of epi models under agentic verification and feedback. Subsequent feedback corrects structural violations and progressively refines model expressiveness.}
    \label{fig:itr}
\end{figure}
\vspace{-1em}
\begin{figure}[h]
    \centering
    \includegraphics[width=\linewidth]{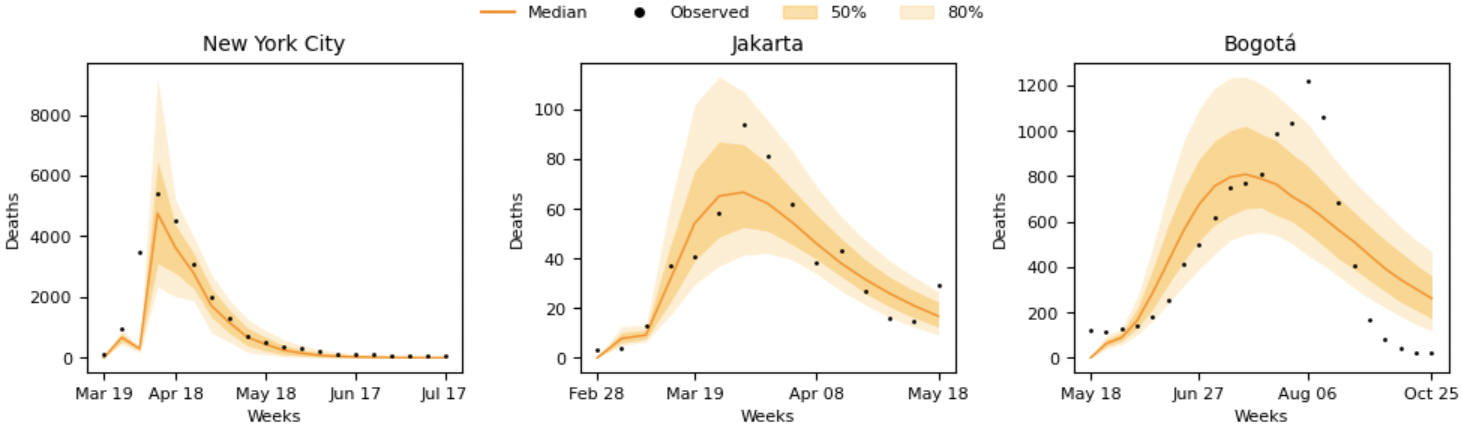}
    \caption{Fitted curves of weekly COVID-19 deaths (counts) across the three geographies considered, using the Data-Driven Behavioral epidemiological model instantiated by \tool{}.}
    \label{fig:baselineEpi}
\end{figure}
\noindent
\textbf{Case Study I.} We evaluate six counterfactual scenarios (A–F) spanning two variations: \emph{vaccination policy}—(i) no recommendation with negligible uptake (A, B), (ii) targeted boosters for high-risk populations aged 65+ (C, D), and (iii) universal booster for all eligible groups (E, F)—and \emph{immune escape level} (low vs.\ high). As shown in Figure~\ref{fig:sc17_correct}, the synthesized simulators exhibit strong epidemiological consistency. Importantly, these qualitative behaviors are \emph{not manually encoded}, but are inferred by the agent through scenario-conditioned graph construction and simulator synthesis. Within each policy pair, scenarios with high immune escape (B, D, F) yield higher cumulative infections than low-escape counterparts (A, C, E). Across vaccination policies, increasing coverage induces a monotonic reduction in infection burden ($E < C < A$ and $F < D < B$). 
These trends emerge from the agent-constructed learned model, confirming correct counterfactual execution.

\subsection{Effect of Agentic Feedback on Convergence}
\label{ssec:results-convergence}
We find that iterative agentic feedback significantly accelerates convergence compared to unguided generation. As illustrated in Figure \ref{fig:itr}, subsequent feedback cycles correct structural violations and refine model expressiveness—such as adding waning-immunity pathways or vaccination strata—thereby mimicking a professional expert's workflow (Figure \ref{fig:feedBack}, Appendix).

%% file: baseline_exp.tex
\subsection{Behavioral Epidemic Models Synthesis}
\label{ssec:baseline_epipaper}

\noindent
To evaluate the generality of \tool{}, we conducted a baseline case study, (\emph{Case Study II}) following the Data-Driven Behavioral (DDB) epidemiological model introduced in \cite{gozzi2025comparative}. A DDB model incorporates behavioral changes by leveraging mobility data to capture variations in contact patterns. Using \tool{}, we instantiated the DDB model and applied it across diverse locations.
For each location, the DDB model generated by \tool{} is calibrated to weekly aggregated deaths using reported surveillance data. Probabilistic projections are then obtained by constructing a parametric ensemble around the calibrated parameters. Figure~\ref{fig:baselineEpi} reports the median projected trajectory together with 50\% and 80\% uncertainty intervals, overlaid with observed weekly deaths. Across all three locations, the projections capture key epidemic trends and peak timing, which demonstrates that \tool{} can be directly applied to any canonical epidemiological models to produce calibrated, uncertainty-aware projections. Importantly, this case study shows that \tool{} can serve as a general framework for deploying, calibrating, and analyzing standard behavioral epidemic models.

%% file: conclusion.tex
\section{Conclusion}
We present \tool{}, an end-to-end agentic framework designed to automate the scenario-conditioned epidemic modeling. Our multi-agent verification and validation architecture successfully bridges the gap between abstract public health scenarios and executable, mechanistically sound simulators. Our evaluation through case studies confirms that the framework not only achieves high empirical accuracy but also maintains structural integrity and logical consistency across complex counterfactual interventions. 
While effective, the current system relies on fixed optimization and bounded iterative refinement, which may occasionally lead to late-stage regression due to compounding noise. Future work will investigate convergence-aware calibration and adaptive stopping criteria to improve robustness. Overall, \tool{} demonstrates the potential of agentic systems to emulate expert epidemiological workflows and accelerate reliable decision support in evolving public health settings.

%% file: impactStatement.tex
\section*{Impact Statement}
This work advances machine learning for scenario-conditioned epidemiological modeling to support public health analysis. The framework is intended strictly as a decision support and modeling tool, 
improving scalability and interpretability while enforcing epidemiological validity. We do not anticipate significant ethical or societal risks; our framework does not generate or design pathogens, or provide guidance for laboratory experimentation.

%% file: appendix.tex
\section{Overview of \tool{}}
Algorithm~\ref{alg:agentic-epi} summarizes the end-to-end workflow of \tool{}, specifying inputs, outputs, and the sequence of agentic generation steps, while Algorithm~\ref{alg:vv_agent} focuses on the verification and validation of the generated epidemiological models, which are crucial components of our framework.

\input{algorithm}

Table~\ref{tab:notation} lists the notation used throughout the paper.
\input{notation}

\begin{figure*}[h!]
    \centering
    \includegraphics[width=\linewidth]{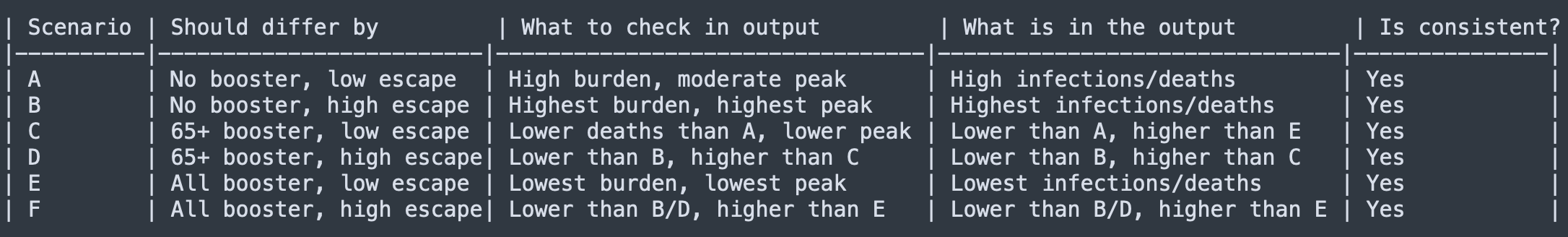}
    \caption{Validation Agent generated scenario-driven differences and validation checks.}
    \label{fig:val_agent}
\end{figure*}

\section{Extended Ablation Study}

\begin{figure*}
    \centering
    \includegraphics[width=\linewidth]{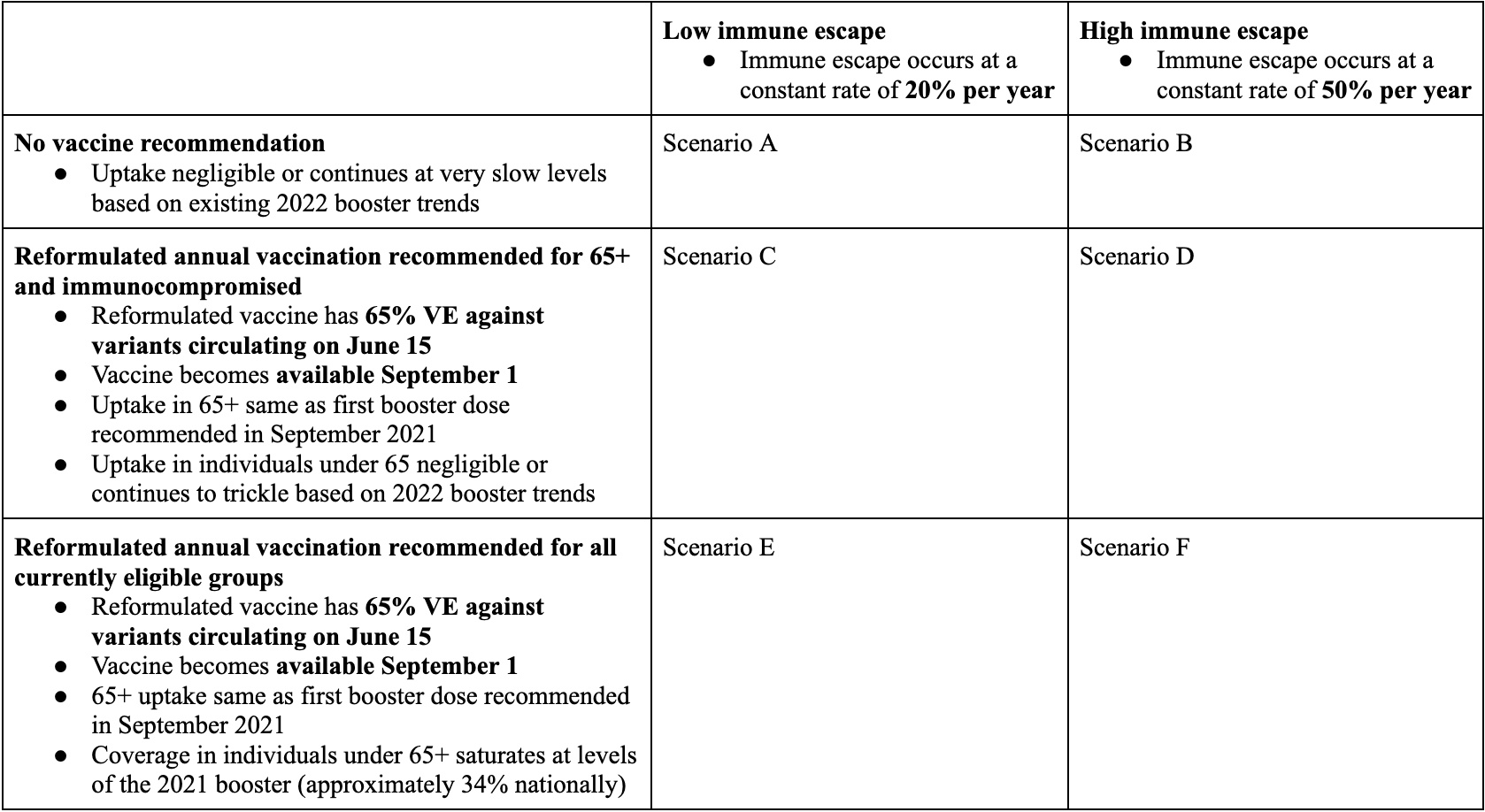}
    \caption{Covid19 Scenario modeling: round 17}
    \label{fig:Covid19Sc17_description}
\end{figure*}
\begin{figure*}
    \centering
    \includegraphics[width=0.8\linewidth]{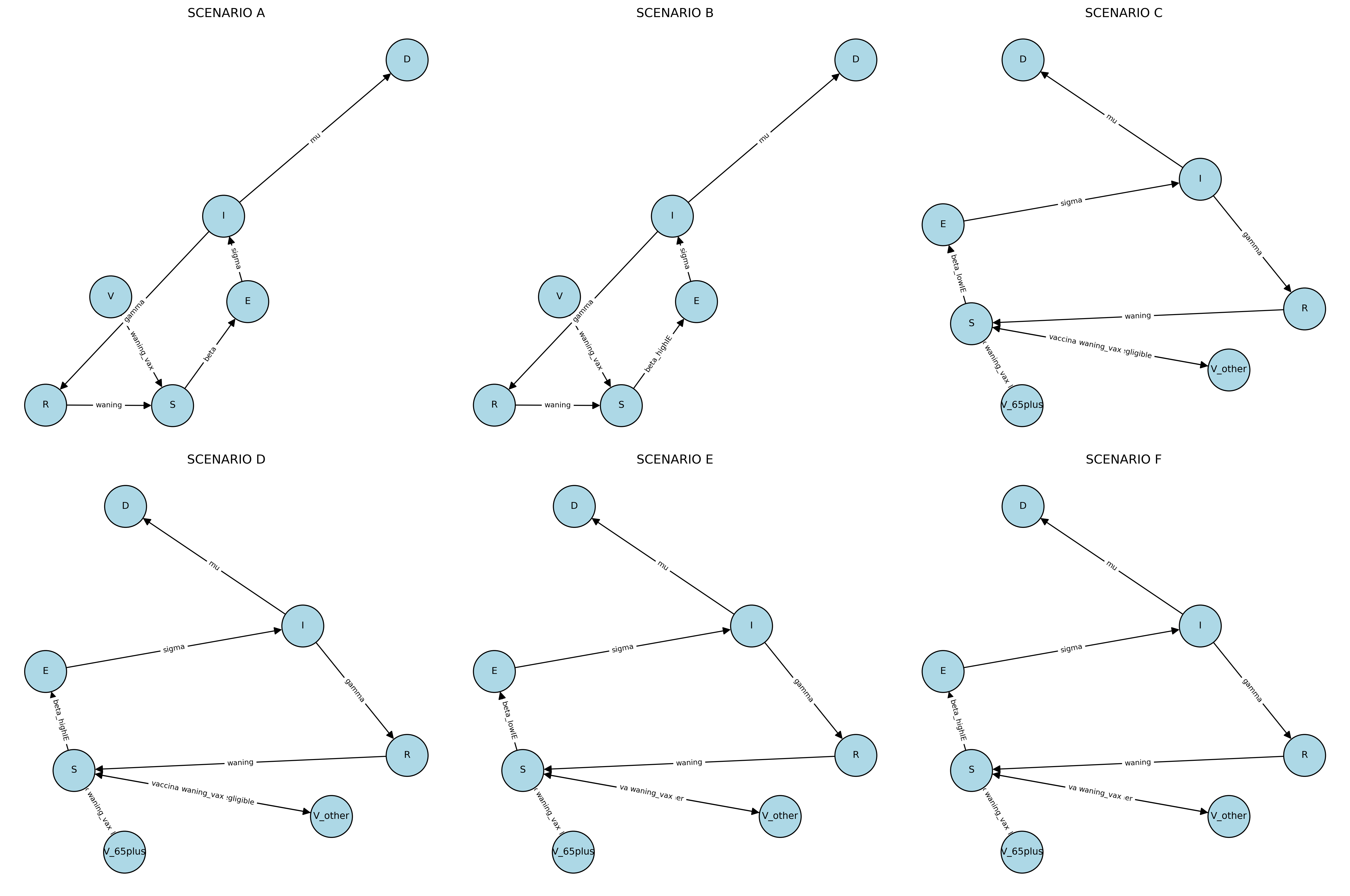}
    \caption{Scenario-based epidemiological flow graph construction. \textit{Retrieved domain knowledge constrains graph structure}, enabling the LLM planner to generate compartmental models.}
    \label{fig:promptToGraph}
\end{figure*}

\begin{figure*}
    \centering
    \includegraphics[width=0.8\linewidth]{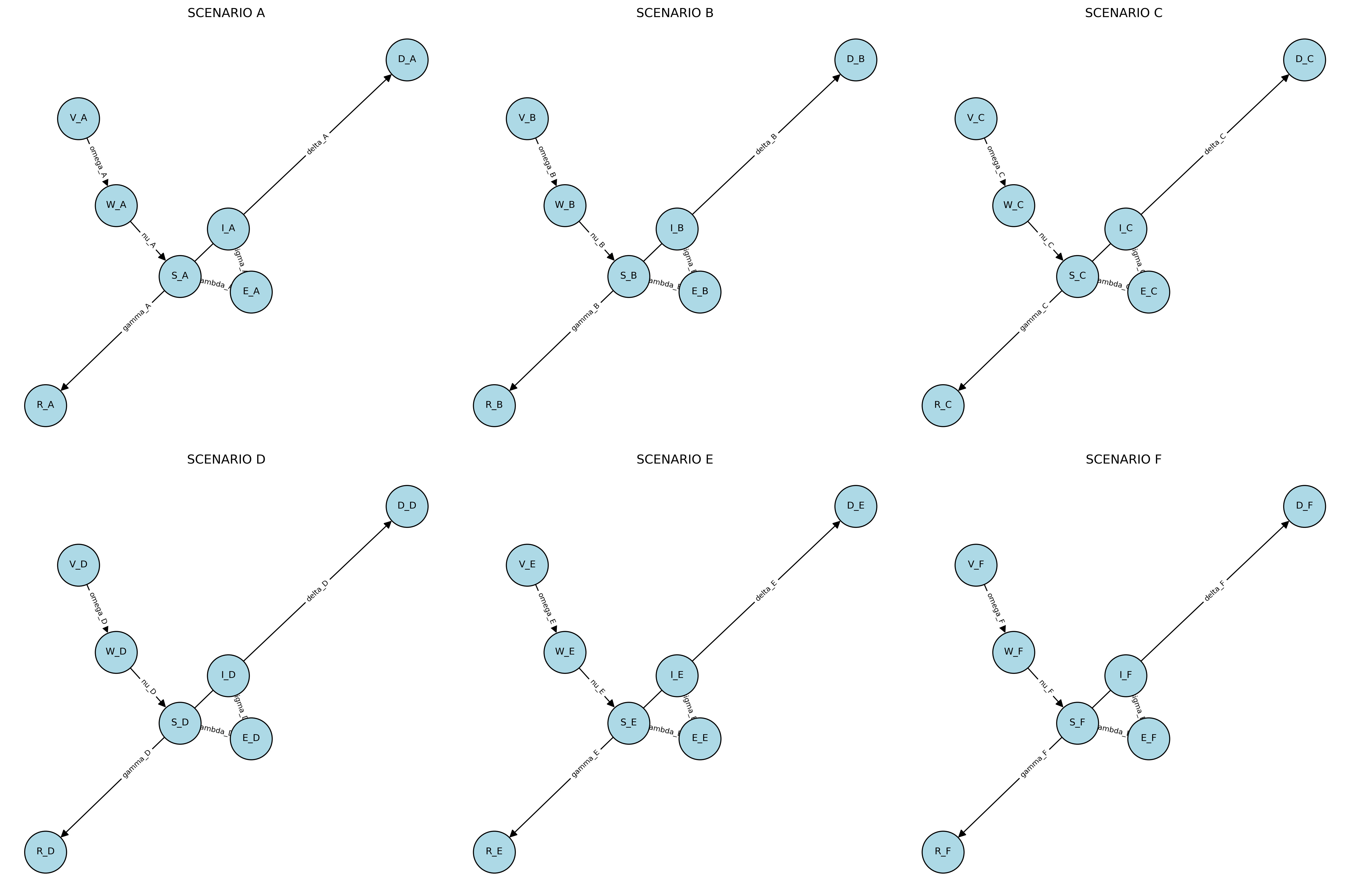}
    \caption{Scenario-based epidemiological flow graph construction \textbf{without external knowledge}, which leads to \textit{oversimplication} and generates almost the \textit{same graphs for all scenarios}}
    \label{fig:promptToGraphwoknowledge}
\end{figure*}

\begin{figure}
    \centering
    \includegraphics[width=0.95\linewidth]{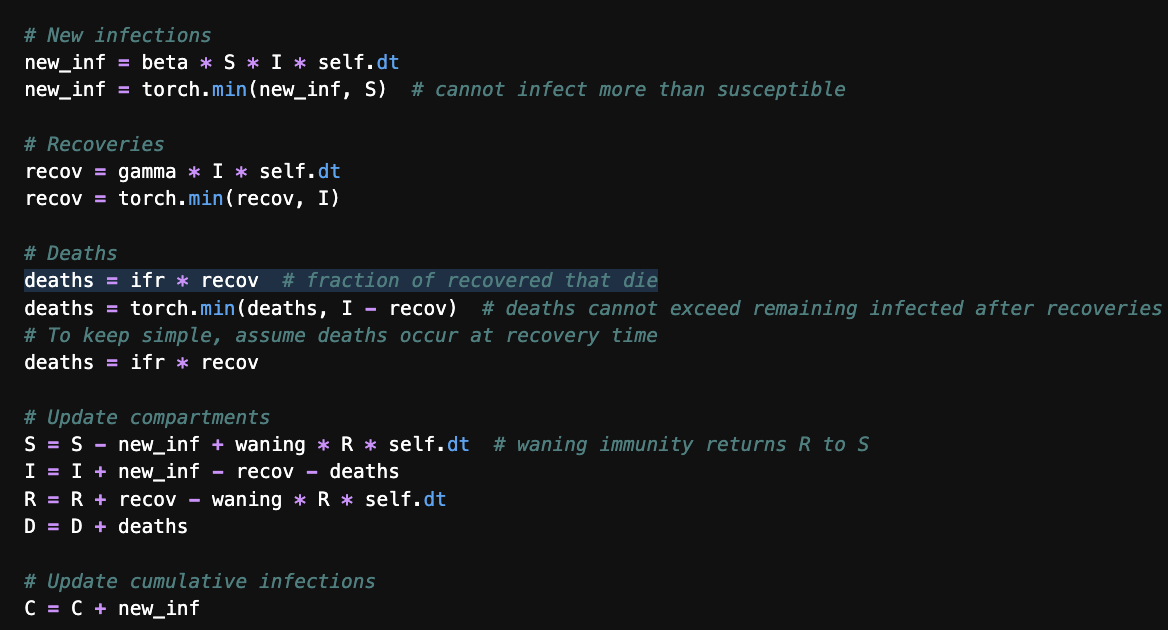}
    \caption{Simulator well fitted, but mechanistic equation invalid. Death is not a fraction of recovered}
    \label{fig:wrongCode}
\end{figure}

\begin{figure*}
    \centering
    \includegraphics[width=0.8\linewidth]{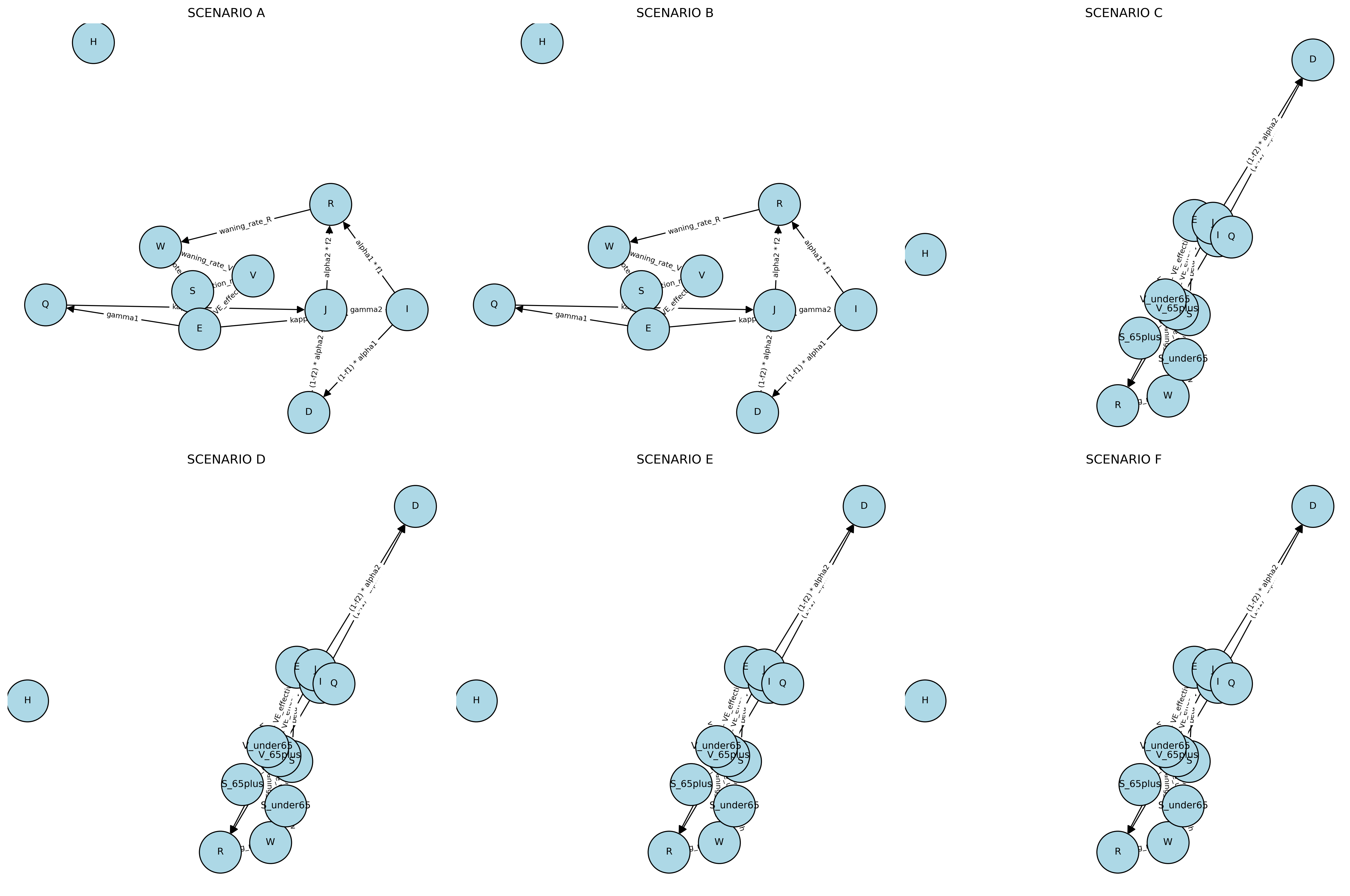}
    \caption{If \textbf{no constraints} are written, the graph generation model will generate the \textit{most complicated flow dynamics}, introducing unnecessary states which are not relevant to the scenarios}
    \label{fig:complicatedGraph}
\end{figure*}

\begin{table*}[t]
\centering
\caption{Impact of language model capacity, structural priors, and constraints on the quality
of automatically generated epidemiological simulators.}
\small
\begin{tabular}{p{3.2cm} p{3.2cm} p{3.2cm} p{3.6cm}}
\toprule
\textbf{Factor} &
\textbf{High-Capacity Models (GPT-4.x)} &
\textbf{Lightweight Models (Mini)} &
\textbf{Key Implications} \\
\midrule

Language model capacity
&
Strong inductive reasoning; infers structure with limited guidance
&
Limited implicit reasoning; sensitive to prompt formulation
&
Higher-capacity models reduce but do not eliminate the need for explicit structure \\

\addlinespace

Skeleton code availability
&
Helpful but not strictly required; improves stability and consistency
&
Critical for correctness; prevents structural and numerical errors
&
Skeleton code acts as a structural prior, especially important for smaller models \\

\addlinespace

Amount of guidance required
&
Low to moderate (high-level objectives often sufficient)
&
High (explicit constraints, parameter bounds, and templates needed)
&
Guidance compensates for reduced model capacity \\

\addlinespace

Behavior without constraints
&
Produces plausible but epidemiologically invalid or unstable dynamics
&
Frequently generates incoherent or numerically unstable simulators
&
Constraints are essential for epidemiological realism across all models \\

\addlinespace

Overall simulator quality
&
High when combined with constraints and minimal scaffolding
&
Competitive only with strong scaffolding and constraint enforcement
&
Reliable model generation emerges from the interaction of all factors \\

\bottomrule
\end{tabular}

\label{tab:model_generation_factors}
\end{table*}

\begin{figure}[htbp]
    \centering
    \includegraphics[width=\linewidth]{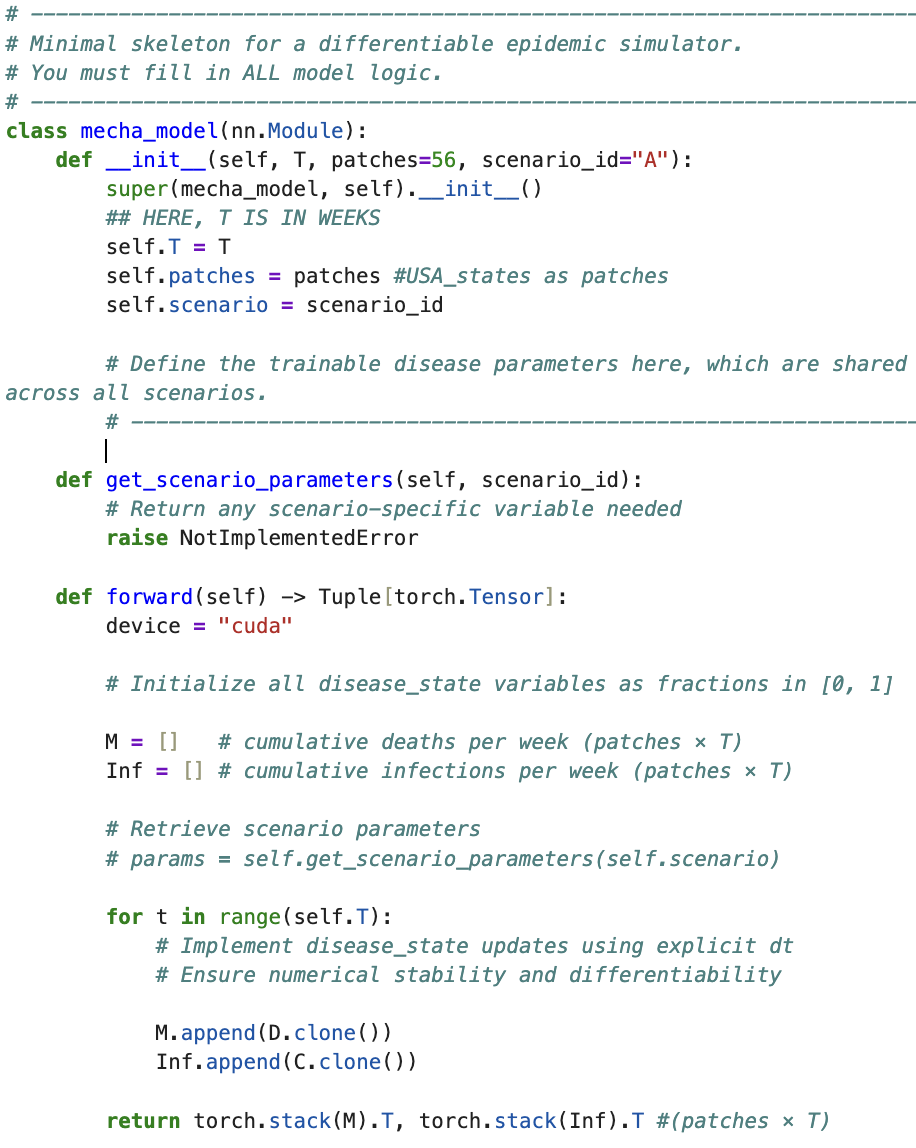}
    \caption{Skeleton Code for Case Study I}
    \label{fig:skeletonCode}
\end{figure}

\begin{figure*}
    \centering
    \includegraphics[width=\linewidth]{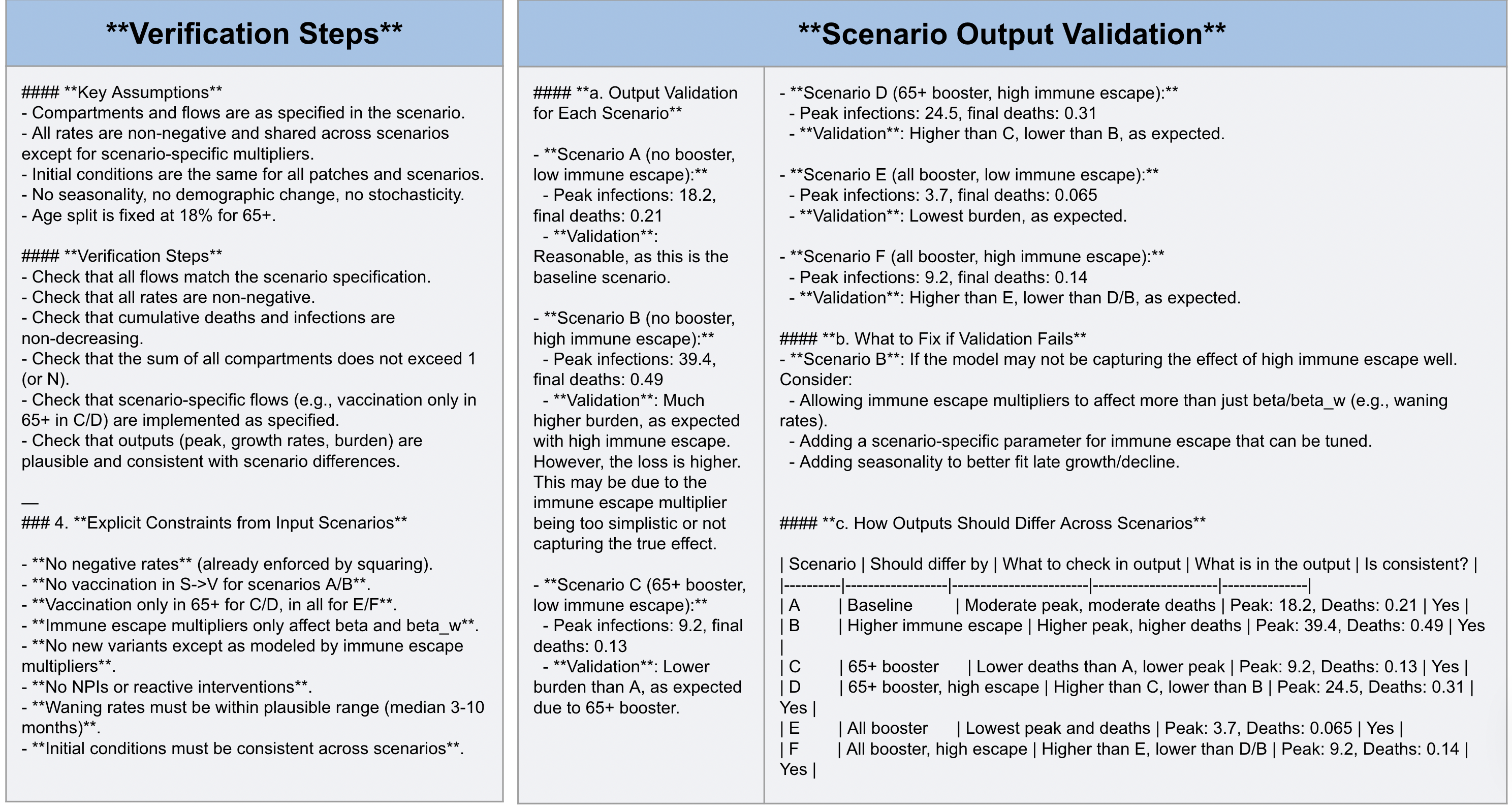}
    \caption{Verification and scenario-output validation in \tool{}. Left: Explicit verification steps and hard constraints enforced on synthesized epidemiological simulators, including mechanistic validity, parameter constraints, and scenario compliance. Right: Scenario-wise output validation illustrating expected qualitative ordering of peak infections and cumulative deaths across counterfactual scenarios, along with diagnostic guidance when validation fails.}
    \label{fig:vnvAgent}
\end{figure*}

\begin{figure*}
    \centering
    \includegraphics[width=0.9\linewidth]{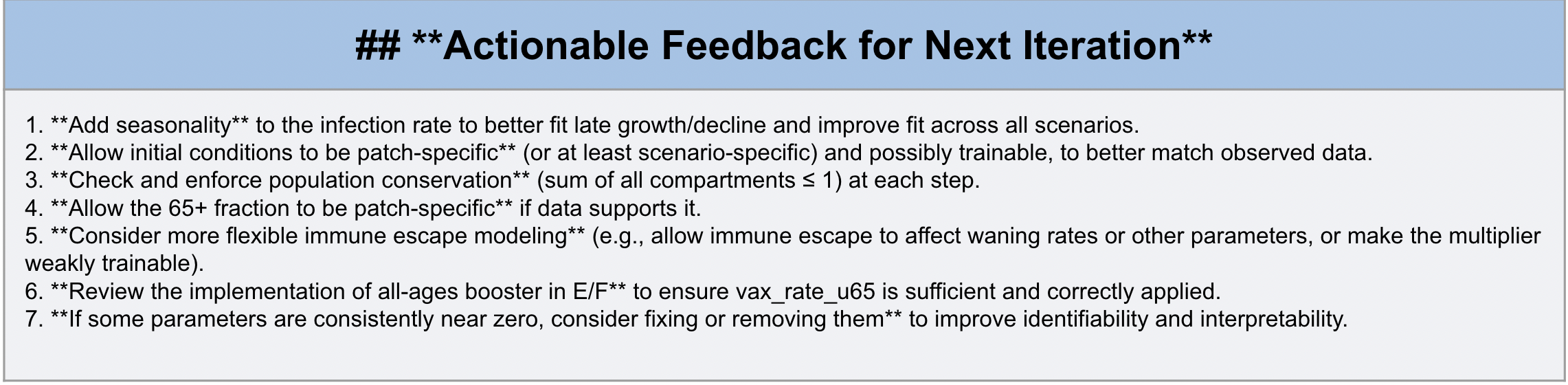}
    \caption{Actionable feedback generated by the agentic refinement loop in model generation phase.}
    \label{fig:feedBack}
\end{figure*}

\begin{algorithm}[h]
\caption{Iterative Verification and Validation (V\&V) Agent}
\label{alg:vv_agent}
\begin{algorithmic}[1]
\REQUIRE Current Program $P_{k}$, Scenarios $\mathcal{S}$, Observed Data $\mathcal{D}$, Error Tolerance $\epsilon$
\ENSURE Refined Model Structure $M_{k+1}$, Validation Status $V \in \{\text{PASS, FAIL}\}$

\WHILE{$\mathcal{L}_{val} > \epsilon$ \AND $k < k_{max}$}
    \STATE \COMMENT{\textbf{Phase I: Structural Verification}}
    \STATE Audit $P_{k}$ for adherence to \textit{Brauer Compartmental Axioms}:
    \STATE \quad 1. \textbf{Non-negativity:} $\forall t, x_i(t) \ge 0$.
    \STATE \quad 2. \textbf{Mass Conservation:} $\sum_{i} \frac{dx_i}{dt} = 0$.
    
    \STATE \COMMENT{\textbf{Phase II: Scenario-Consistency Validation}}
    \FOR{each scenario $s \in \mathcal{S}$}
        \STATE Simulate $\hat{y}_s \leftarrow \text{Exec}(P_k, s)$
        \STATE Evaluate $\Delta_{peak}, \Delta_{rate}$ against scenario-specific constraints $\mathcal{C}_s$.
        \IF{$\text{Sign}(\Delta \hat{y}_s) \neq \text{Sign}(\Delta \mathcal{C}_s)$}
            \STATE $\mathcal{F}_{log} \gets \text{Flag contradiction in } (V, E) \text{ transition logic}$.
        \ENDIF
    \ENDFOR
    
    \STATE \COMMENT{\textbf{Phase III: Agentic Reasoning}}
    \STATE Compute Calibration Residuals: $\mathcal{R} = \| \hat{y} - \mathcal{D} \|$
    
    \IF{$\text{Trend}(\mathcal{R}) \text{ is stalled} > \tau$ \AND $P_k \in \text{White-Box}$}
        \STATE $M_{k+1} \gets P_k + \mathcal{N}_{\phi}(t, x)$\\
        \COMMENT{Inject Neural Residual Terms, If specified in prompt}
        \STATE Update prompt context: $\mathcal{P} \gets \mathcal{P} \oplus \text{"limitations reached; initiating hybrid calibration."}$
    \ELSE
        \STATE \COMMENT{Maintain mechanistic interpretability but refine parameters}
        \STATE $M_{k+1} \gets \text{RefactorLogic}(P_k, \nabla_{\theta}\mathcal{L}_{val}, \mathcal{F}_{log})$
    \ENDIF
    
    \STATE $k \gets k + 1$
\ENDWHILE
\STATE RETURN $M_{k}$
\end{algorithmic}
\end{algorithm}

\begin{table}[]
\centering
\caption{Unified hard constraints enforced during flow-graph verification and simulator synthesis. 
Here $*$ denotes any compartment.}
\small
\setlength{\tabcolsep}{4pt}
\begin{tabular}{p{2.8cm} p{5.4cm}}
\toprule
\textbf{Constraint Type} & \textbf{Specification} \\
\midrule
\textbf{Flow Validity} &
$S \rightarrow R$ (no direct recovery) \\
& $S \rightarrow I$ (latent period required) \\
& $\{S,E,R,V,W\} \rightarrow D$ (death only from severe states) \\
& $D \rightarrow *$ forbidden (terminal state) \\
& $\{E,I,J,H\} \rightarrow V$ (vaccination eligibility) \\
& $\{S,E,I,J,H\} \rightarrow W$ (waning semantics) \\
& $\{E,I,R,J,H\} \rightarrow E$ (exposure semantics) \\

\midrule
\textbf{Execution Interface} &
Must complete a fixed PyTorch skeleton \\

\textbf{Differentiability} &
End-to-end differentiable; no parameter reassignment in \texttt{forward()} \\

\textbf{Numerical Stability} &
Explicit time steps; no NaNs or Infs \\

\textbf{State Semantics} &
All states represent population fractions in $[0,1]$ \\

\textbf{Output} &
Return cumulative infections and deaths ($P \times T$) \\

\textbf{Parameter Sharing} &
Differentiable disease parameters are shared across all scenarios \\

\textbf{Scenario Isolation} &
Scenario-specific variables are non-trainable \\

\textbf{Constraint Handling} &
No forced non-negativity (e.g., ReLU/Softplus) \\
\bottomrule
\end{tabular}
\label{tab:unified_constraints}
\end{table}

\begin{figure*}
    \centering
    \includegraphics[width=0.95\linewidth]{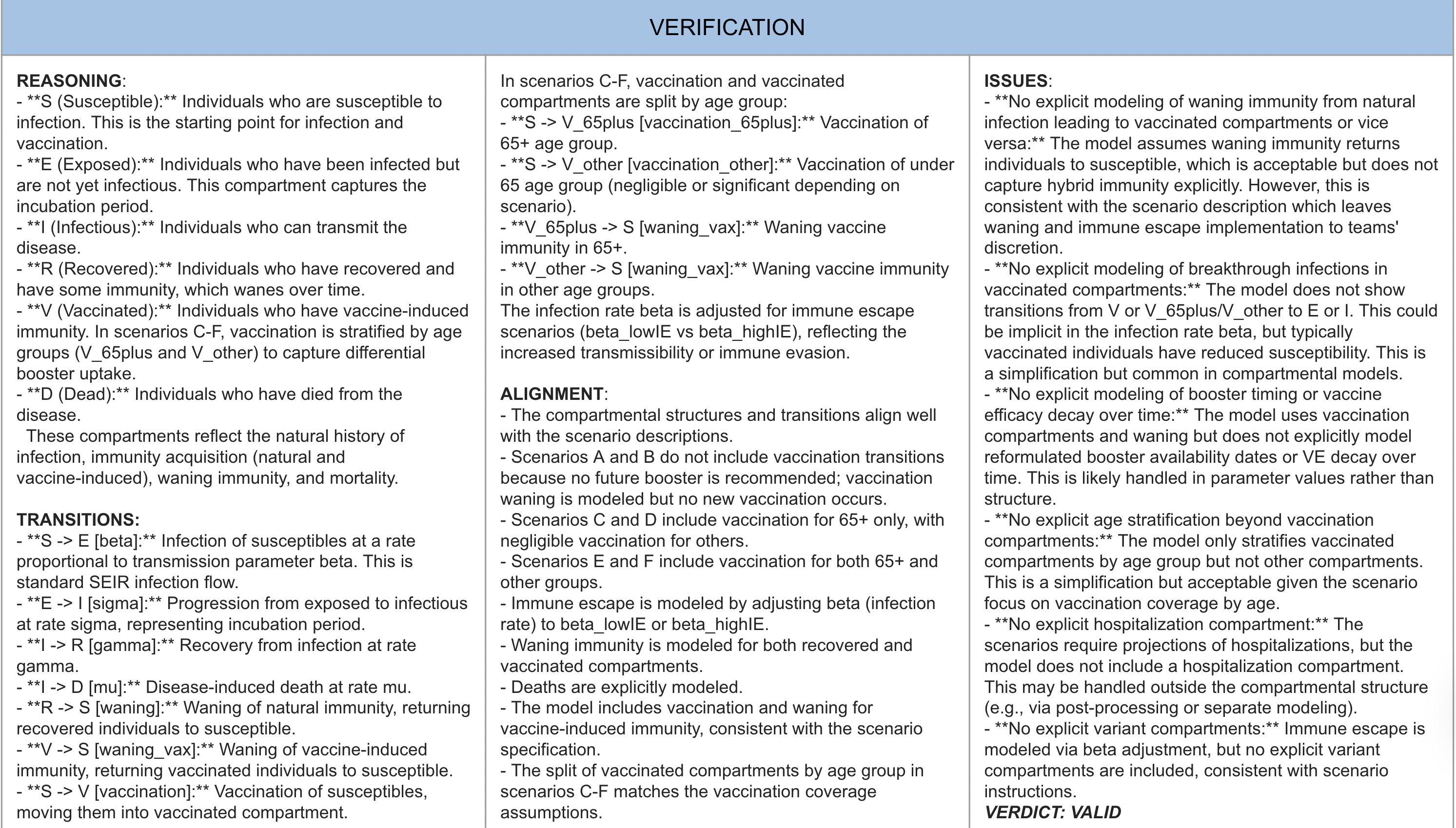}
    \caption{Graph Verification checks and reasoning while generating the flow to ensure consistency between the scenarios}
    \label{fig:graphVerfication}
\end{figure*}

\paragraph{Without Flow-Graph Verification}
When the verification step is removed, the LLM frequently produces epidemiologically invalid transitions, such as direct $V->E$ flows, which violate basic disease progression semantics. Incorporating a verification layer that enforces structural constraints (e.g., valid infection pathways and intervention eligibility) is therefore necessary to ensure the correctness of the generated flow graphs. Figure \ref{fig:graph_error_propagation} illustrates how errors in graphs may propagate and lead to incorrect solutions.

\paragraph{Without External Knowledge Retrieval (No RAG)}
Removing external epidemiological knowledge retrieval results in simpler flow structures, typically defaulting to canonical SEIRD-style models without explicit quarantine or isolation compartments. In contrast, when domain knowledge is retrieved and provided to the LLM, the generated graphs consistently include epidemiologically meaningful compartments and transitions (e.g., exposed, quarantined, isolated), aligned with the given scenarios. An additional observation is that with knowledge augmentation, the generated graphs often satisfy structural constraints in the first iteration, whereas without retrieval, multiple generate--verify cycles are required for the LLM to converge to a valid flow.

\paragraph{Without Flow-Graph Intermediate Representation (Direct Scenario-to-Code)}
In this ablation, the LLM is prompted to generate executable model equations directly from the scenario description, bypassing the intermediate flow-graph representation. We observe that this approach leads to higher rates of structural errors in the generated equations, including missing compartments and invalid transitions (i.e., $R->D$), which are harder to detect from the code. Introducing an explicit flow-graph intermediate representation improves reliability by decoupling structural validation from equation generation and enabling targeted verification before code synthesis.

%% file: algorithm.tex
\begin{algorithm}
\caption{Agentic Epidemiological Model Generation}
\label{alg:agentic-epi}
\begin{algorithmic}[1]
\REQUIRE Scenario $\mathcal{S}$, $\mathcal{D}$, constraints $\mathcal{C}$,code skeleton $\mathcal{CK}$
\ENSURE Validated, scenario-consistent simulator, $f_\theta^{\mathcal{S}}(\cdot)$, Optimized parameters $\theta^*$, Reasoning $\mathcal{R}$

\STATE \COMMENT{\textbf{Stage I: Knowledge-Augmented Graph Synthesis}}
\STATE $\mathcal{K} \leftarrow \textsc{RAG}(\{\mathcal{S}\}, \text{EpiKnowledgeBase})$ 
\STATE $\mathcal{P}_{graph} \leftarrow \textsc{Prompt}(\{{s}_i\}, \mathcal{C}, \mathcal{K})$

\REPEAT
    \STATE Generate flow graph $\mathcal{G}_g \gets \mathcal{A}_{graph}( \mathcal{P}_{graph})$ 
    \STATE $\mathcal{V}_{graph} \leftarrow \textsc{Graph\_Verification}(\mathcal{G}_g, \mathcal{C})$ \\
    \COMMENT{Check transitions \& Agentic Verification}
    \STATE $\mathcal{P}_{graph} \leftarrow \mathcal{P}_{graph} \oplus \mathcal{A}_{Feedback}(\mathcal{V}_{graph})$
\UNTIL{$\mathcal{V}_{graph} = \textsc{Pass}$}

\STATE \COMMENT{\textbf{Stage II: Iterative Functional Implementation}}
\STATE $\mathcal{P}_{code} \leftarrow \textsc{Prompt}(\mathcal{G}_g, \mathcal{CK}, \mathcal{S})$
    
\FOR{generation $g = 1,\ldots,G$}
    \STATE Mechanistic Simulator $\mathcal{C}_g \gets \mathcal{A}_{planner}( \mathcal{P}_{code}, \mathcal{G}_g)$ \\
     \IF{$\mathcal{C}_g$ execution fails}
        \STATE  $\mathcal{P}_{code} \leftarrow \mathcal{P}_{code} \oplus \textsc{ErrorTrace}(\mathcal{C}_g)$
        \STATE \textbf{continue}
    \ENDIF
    \STATE \COMMENT{\textbf{Stage III: Hybrid Calibration \& V\&V}}    
    \STATE \textbf{Train:} $\theta_g \leftarrow \arg\min_\theta \mathcal{L}(\mathcal{C}_g,\mathcal{D})$ 
    \STATE $\mathcal{R}_g \leftarrow \textsc{get\_StructuralReasoning}(\mathcal{C}_g, \theta_g)$
    \STATE $\mathcal{V}_{g} \leftarrow \textsc{check\_V\&V}(\mathcal{C}_g, \theta_g, \{\mathcal{S}_i\})$ 
    
    \IF{$\mathcal{V}_{g} = \textsc{Pass}$ \AND $\mathcal{L} < \epsilon$}
        \STATE \textbf{return} $(f_{\theta_g}, \theta_g, \mathcal{R}_g)$
    \ELSE
        \STATE $\mathcal{F}_g \leftarrow \textsc{get\_Feedback}(\mathcal{V}_g, \mathcal{L})$ 
        \STATE $\mathcal{P}_{code} \leftarrow \mathcal{P}_{code} \oplus \mathcal{F}_g$
    \ENDIF

\ENDFOR

\textit{Return} Final validated simulator code $\mathcal{C}^*$ and optimized parameters $\theta^*$
\end{algorithmic}
\end{algorithm}

%% file: notation.tex
\begin{table}[t]
\centering
\caption{Summary of notations}
\small
\resizebox{1\linewidth}{!}{%
\begin{tabular}{p{1.6cm}| p{7cm}}
\toprule
\textbf{Symbol} & \textbf{Description} \\
\midrule
${s}_i$ & Scenario specification \\
$\mathcal{D}$ & Observed epidemiological data (infections, deaths) \\
$p$ & patch (spatial or age groups) \\
$t$ & Discrete time index (weeks) \\

$G=(V, E)$ & Epidemiological flow graph with compartments $V$ and transitions $E$ \\
$x \in V$ & Disease compartment (e.g., $S, E, I, R, V, W$) \\
$(u,v) \in E$ & Directed transition from compartment $u$ to $v$ with weight of disease parameter \\
$\mathcal{P}_{graph}, \mathcal{P}_{code}$, & Prompt to generate flow graph \& code\\
$\mathcal{C}_g$ & Generated executable simulator corresponding to $G$ \\
${\theta^*}$ & Calibrated parameter estimates after training \\
$f^{s}_{\theta^*}(\cdot)$ & Epidemiological simulator \\
$\mathcal{L}(\cdot)$ & MSE loss over observed trajectories \\
$\mathcal{A}_{\text{plan}}$ & LLM-based planner agent for code generation\\
$\mathcal{K}$ & Retrieved epidemiological knowledge corpus (RAG) \\
${RAG}(\cdot)$ & FAISS-based retrieval operator \\
$\mathcal{C}$ & Set of epidemiological and scenario constraints \\
$\mathcal{CK}$ & Skeleton Code for LLM Planner Agent \\
\bottomrule
\end{tabular}
}
\label{tab:notation}
\end{table}

%% file: agent_table.tex
\section{Role of Agents in \tool{}}

\tool{} relies on a collection of specialized agents, each responsible for a distinct stage of the epidemic simulator construction and validation pipeline. Together, these agents ensure that the generated models are structurally correct, epidemiologically meaningful, and executable.

In particular, agents are used for the following tasks:
\begin{enumerate}
    \item \textbf{Flow-graph construction.}  
    Given a natural-language scenario description and retrieved epidemiological knowledge, an agent constructs a compartmental flow graph representing disease states and admissible transitions. This graph encodes the high-level mechanistic structure implied by the scenario, such as vaccination policies, waning immunity, immune escape, and behavioral effects.

    \item \textbf{Flow-graph verification.}  
    The constructed flow graph is evaluated by a dedicated verification agent. This agent enforces hard epidemiological constraints (e.g., valid causal transitions, conservation of population mass, and absence of impossible flows such as direct $S \rightarrow R$ recovery). Graphs that violate these constraints are rejected, and structured error feedback is returned to guide regeneration.

    \item \textbf{Model instantiation and calibration.}  
    Once a flow graph is verified, it is passed to an LLM-based planner agent, which translates the graph into a system of ordinary differential equations and corresponding executable simulator code. The resulting simulator is then calibrated against real-world epidemiological data using gradient-based optimization or sampling-based inference.

    \item \textbf{Code-level verification and validation.}  
    The generated simulator code is further analyzed by a set of agents responsible for execution-time verification and validation. These agents check for numerical instability, invalid state evolution (e.g., negative compartment values), violations of population conservation, and non-monotonic cumulative quantities. Feedback from these agents is used to iteratively correct the code when necessary.
\end{enumerate}

Through this multi-agent design, \tool{} enforces correctness at multiple levels: structural (flow graph), dynamical (ODE system), and executable (simulation behavior). The distinct roles of these agents, including the LLM-based planner, are summarized in Table~\ref{tab:vnv_agents_roles}.

\begin{table*}[t]
\centering
\caption{Agent responsibilities and guarantees in the VnV-driven generation framework. Each agent enforces a distinct modeling aspect—ranging from mathematical correctness to scenario plausibility—ensuring that successive generations improve structural validity, interpretability, and scenario fidelity rather than only minimizing validation loss.}\small
\begin{tabular}{p{1.2cm} p{3.4cm} p{1.8cm} p{3.8cm} p{4.8cm}}
\toprule
\textbf{Agent} &
\textbf{Primary Function} &
\textbf{Aspect} &
\textbf{Outputs Produced} &
\textbf{How This Improves Model Generation} \\
\midrule

\textbf{Verification Agent} &
Evaluates hard mathematical and epidemiological constraints on simulator states and parameters &
\textbf{Correctness} \newline
(legal dynamics, numerical validity) &
Binary verdict (\texttt{PASS}/\texttt{FAIL}); violation-specific error logs (e.g., negative rates, mass imbalance) &
Prevents invalid or unphysical models from entering optimization or interpretation stages; constrains the search space to epidemiologically admissible simulators \\

\midrule

\textbf{Validation Agent} &
Assesses whether verified outputs are scientifically meaningful and consistent with scenario intent &
\textbf{Plausibility \& Scenario Fidelity} \newline
(realistic dynamics, intervention effects) &
Scenario-level validation status (\texttt{PASS}/\texttt{WARN}/\texttt{FAIL}); diagnostics such as scenario collapse or missing mechanisms &
Detects structurally insufficient models that achieve low loss but fail to express intended scenario differences; forces mechanistic adequacy beyond curve fitting \\

\midrule

\textbf{Reasoning Agent} &
Interprets logs and optimized parameters to identify structural causes of failure &
\textbf{Interpretability \& Diagnosis} &
Mechanistic explanations linking observed failures to missing compartments, flows, or constraints &
Prevents repeated generation of equivalent white-box models; explains why loss minimization succeeded or failed in epidemiological terms \\

\midrule

\textbf{Feedback Agent} &
Converts reasoning diagnostics into concrete, actionable regeneration directives &
\textbf{Learnability Across Generations} &
Explicit structural modification instructions (e.g., add vaccination compartments, separate waning pathways) &
Guides the LLM planner toward targeted architectural changes, accelerating convergence toward valid and expressive models \\

\midrule

\textbf{LLM Planner Agent} &
Synthesizes feedback with prior generation history to design the next simulator specification &
\textbf{Structured Exploration} &
Revised model blueprint specifying compartments, parameters, and scenario logic &
Ensures iterative improvement occurs at the model-structure level, not merely via parameter re-optimization \\

\bottomrule
\end{tabular}

\label{tab:vnv_agents_roles}
\end{table*}

In Figure \ref{fig:vnvAgent} and ~\ref{fig:feedBack}, we illustrate representative examples of the checks performed during the verification process. These examples highlight the types of structural and execution-level constraints enforced by the agents; however, the verification is not limited to these cases, as the exact checks invoked may vary due to stochastic generation and scenario-specific model structure.

%% file: agent_result_table.tex
\begin{table*}[h]
\centering
\caption{Verification, validation, and feedback assessment of the final generated epidemic simulator. Verification enforces hard correctness constraints, validation ensures scientific plausibility and scenario fidelity, and feedback identifies concrete structural improvements for subsequent model generations.}
\resizebox{1\linewidth}{!}{%
\begin{tabular}{p{1.8cm}| p{9.5cm}| p{2.2cm}| p{9cm}}
\toprule
\textbf{Agent} &
\textbf{What Was Checked} &
\textbf{Status} &
\textbf{Implication / Actionable Outcome} \\
\midrule
\textbf{Verification} &
\begin{itemize}\setlength{\itemsep}{0pt}
\vspace{-1.5em}
\item Non-negativity of all compartments ($S,E,I,R_{\text{nat}},R_{\text{vac}},D,C$)
  \item Conservation of population mass (excluding deaths)
  \item Monotonicity of cumulative infections and deaths
  \item Non-negativity of disease parameters ($\beta,\gamma,\mu,\sigma,\text{waning}$)
  \item Numerical stability across all scenarios
\end{itemize}
&
\textbf{PASS}
&
Confirms epidemiological correctness and mathematical validity; prevents invalid or unphysical simulators from being considered regardless of validation loss \\

\midrule

\textbf{Validation} &
\vspace{-1.5em}
\begin{itemize}\setlength{\itemsep}{0pt}
  \item Scenario fidelity across A--F (immune escape, booster coverage)
  \item Cross-scenario ordering of infections and deaths
  \item Presence of meaningful intervention effects (boosters reduce burden)
  \item Relative severity of high vs.\ low immune escape scenarios
\end{itemize}
&
\textbf{PASS}
&
Demonstrates scientific plausibility and correct expression of scenario mechanisms beyond curve fitting; confirms that scenarios differ for epidemiologically meaningful reasons \\

\midrule

\textbf{Feedback} &
\vspace{-1.5em}
\begin{itemize}\setlength{\itemsep}{0pt}
  \item Sensitivity of effective transmission ($R_t$) to scenario mechanisms
  \item Identification of missing couplings (vaccination, immune escape, transmission)
  \item Assessment of structural limitations despite correct outputs
\end{itemize}
&
\textbf{ACTIONABLE}
&
Triggers targeted structural improvements for the next generation, including gradual vaccination uptake, waning vaccine efficacy, age-specific mortality, seasonal forcing, and scenario-dependent transmission \\

\midrule

\textbf{Decision} &
Joint assessment of correctness, plausibility, and scenario expressiveness using VnV and feedback agents &
\textbf{ACCEPTED}
&
Model is epidemiologically valid and scenario-consistent, while VnV-driven feedback provides clear guidance for extending expressiveness and improving realism in subsequent generations \\
\bottomrule
\end{tabular}
}

\label{tab:vnv_feedback_summary}
\end{table*}

%% file: discussion.tex
\section{Discussion}
\label{sec:discussion}
\paragraph{Impact of Language Model Choice.}
Our results indicate that the quality and reliability of the generated epidemiological
simulators are strongly influenced by the choice of underlying language model. Larger and more capable models, e.g., GPT-4.1, in comparison to GPT-4.1 mini, GPT-4-0, GPT-4-0-mini, consistently produce simulator code that better adheres to epidemiological structure and scenario semantics, even under
relatively weak prompting. In contrast, smaller or lightweight models rely on the skeleton code and are more prone to numerical instability or violations of epidemiological constraints when operating without explicit guidance. We summarize this observation in Table \ref{tab:model_generation_factors} and \ref{tab:generation_factors} .

\paragraph{Role of Skeleton Code and Structural Priors.}
Across all model variants, the presence of a skeleton codebase improves generation quality.
Skeleton code acts as a strong structural prior that constrains the search space of possible programs, reducing ambiguity in both model dynamics and data flow. While larger models are sometimes able to infer missing structure implicitly, smaller models rely heavily on such explicit framing to avoid degenerate or incoherent simulator implementations. This suggests that skeleton code and language model capacity play complementary roles in simulator generation. We show a sample skeleton code for Case Study I in Figure \ref{fig:skeletonCode}.

\paragraph{Effect of Constraints on Generated Simulators.}
When constraints are removed or weakened, all model variants exhibit degraded performance,
though the failure modes differ. Larger models tend to violate epidemiological realism
(e.g., implausible parameter values or unstable long-term dynamics), while smaller models
more frequently produce structurally invalid simulators or numerically unstable updates.
Explicit constraint enforcement is therefore essential not only for epidemiological
correctness but also for maintaining robustness across different language model capacities.

Table~\ref{tab:unified_constraints} summarizes the hard structural and execution constraints enforced throughout model synthesis. These constraints restrict allowable compartmental transitions to epidemiologically valid flows (e.g., enforcing latent infection, terminal death states, and semantically correct vaccination and waning pathways) while simultaneously constraining simulator execution to a fixed, differentiable PyTorch interface with stable numerical updates. By fixing parameter sharing, scenario isolation, and optimization behavior, the framework ensures that candidate simulators differ only in their structural mechanisms rather than optimizer tuning or re-fitting artifacts.
\begin{table}
\centering
\caption{Effect of language model capacity, skeleton code, and constraints on simulator
generation quality. ``++'': strong, ``+'': moderate, ``--'': poor.}
\resizebox{1\linewidth}{!}{%
\begin{tabular}{p{4.2cm}| p{0.9cm}| p{0.8cm}| p{3cm}}
\toprule
\textbf{Factor} &
\textbf{GPT-4.x} &
\textbf{Mini} &
\textbf{Implication} \\
\midrule
Reasoning ability        & ++ & +  & Capacity matters \\
Skeleton code               & +  & ++ & Structure is critical \\
Guidance                  & +  & ++ & Guidance compensates capacity \\
Stable without constraints        & +  & -- & Constraints essential \\
Epidemiologically realistic output& +  & +  & Depends on scaffolding \\
\midrule
Reliable simulator generation     & ++ & +  & Requires all factors \\
\bottomrule
\end{tabular}
}

\label{tab:generation_factors}
\end{table}
\paragraph{Structural Revision over Parameter Clipping.} Imposing hard constraints, such as $ReLU$ activations on disease parameters, can mask underlying model misspecification. In early iterations, the optimizer frequently drives parameters to negative values when the compartmental logic is incompatible with the ground truth data. By treating these boundary violations as signals for topological revision rather than simple optimization constraints, the VnV agent forces a re-evaluation of the model structure. This ensures that non-negativity is a property of a correct model rather than a forced numerical artifact.

\section{Design Choices for Disease Parameterization}
\label{sec:design-choices}

A key design decision in \tool{} concerns how disease dynamics are parameterized within the generated simulators. We consider three increasingly expressive modeling choices—\emph{time-invariant}, \emph{time-varying}, and \emph{neural-augmented} dynamics—each trading off interpretability, flexibility, and data efficiency.

\paragraph{Time-invariant mechanistic models.}
In the simplest setting, disease parameters are assumed constant over time. Let
$x(t) \in \mathbb{R}_+^K$ denote the compartmental state vector induced by the verified flow graph.
The epidemic dynamics are governed by a system of ordinary differential equations
\[
\frac{d x(t)}{dt} = f_{\text{mecha}}\!\left(x(t); \theta \right),
\]
where $\theta = \{\beta, \sigma, \gamma, \ldots\}$ are fixed disease parameters such as transmission, incubation, and recovery rates.
This formulation corresponds to classical compartmental models and favors interpretability and robustness.
In \tool{}, this option is selected by default unless the scenario specification explicitly requires temporal adaptation.

\paragraph{Time-varying parameter models.}
To capture non-stationary dynamics induced by interventions, behavior change, or seasonal effects, \tool{} optionally allows disease parameters to vary over time.
In this setting, the dynamics take the form
\[
\frac{d x(t)}{dt} = f_{\text{mecha}}\!\left(x(t); \theta(t), \phi(s)\right),
\]
where $\theta(t)$ denotes time-dependent parameters (e.g., $\beta(t)$ or $\gamma(t)$) and $\phi(s)$ encodes scenario-specific controls.
Time variation may be implemented through parametric functions or externally provided covariates (e.g., mobility or seasonality). This design improves short-term fit while preserving the underlying mechanistic structure. In particular, when the observed data exhibit multiple epidemic waves or recurrent peaks, time-varying disease parameterization enables the model to adapt transmission dynamics over time and achieve substantially improved fidelity compared to time-invariant formulations.

\begin{figure*}
    \centering
\includegraphics[width=1\linewidth]{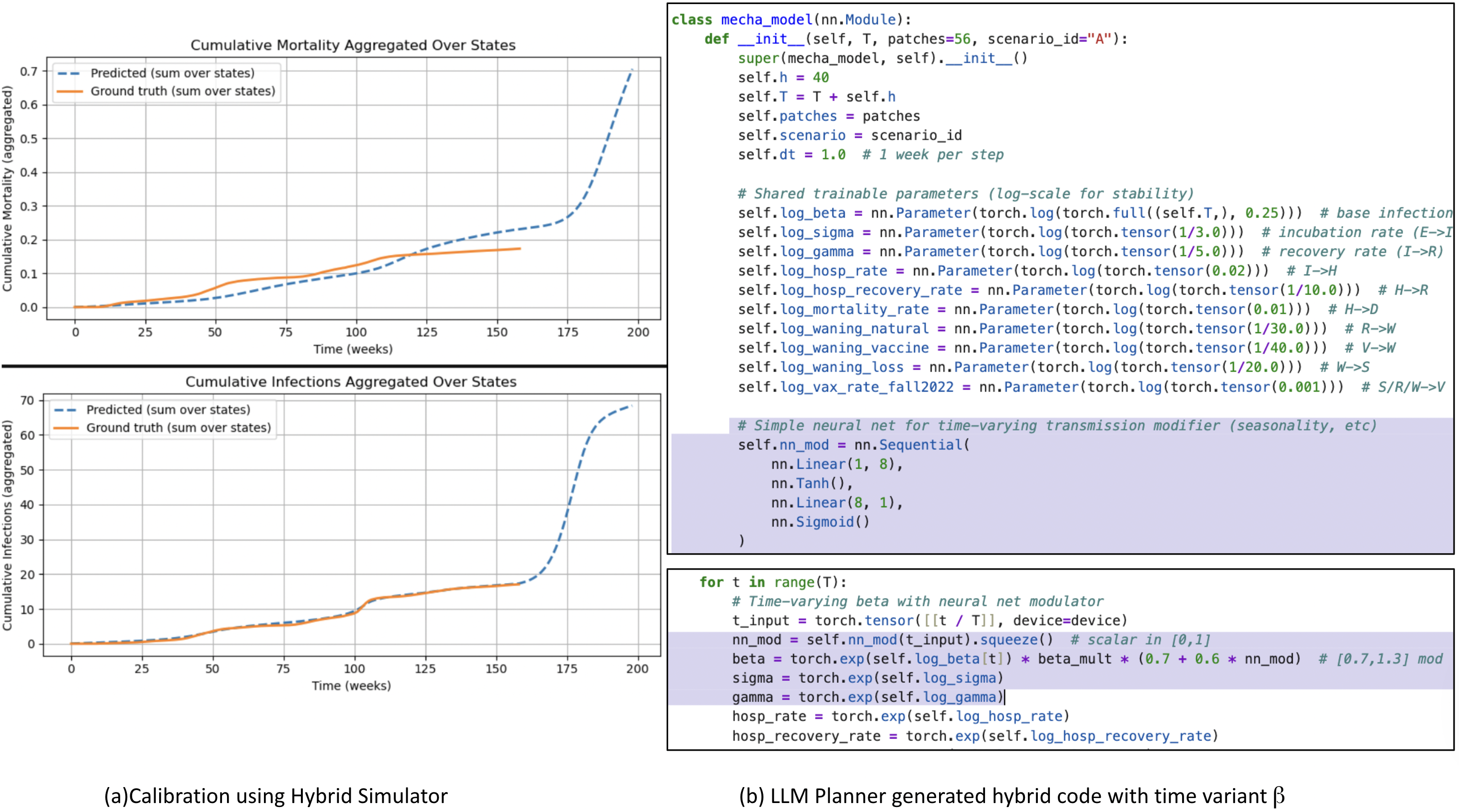}
    \caption{Neural Augmented Dynamics with time variant disease parameters. }
    \label{fig:hybrid}
\end{figure*}

\paragraph{Neural-augmented dynamics.}
When additional flexibility is required, \tool{} supports augmenting mechanistic models with neural components.
Specifically, a neural function $g_\psi(\cdot)$ is introduced to model residual or latent effects:
\[
\frac{d x(t)}{dt} =
f_{\text{mecha}}\!\left(x(t); \theta \right)
\;+\;
g_\psi\!\left(x(t), t\right),
\]
where $g_\psi$ is parameterized by a neural network with parameters $\psi$.
This hybrid formulation allows the simulator to learn complex, time-varying effects while retaining an interpretable mechanistic backbone.
To avoid over-parameterization, neural augmentation is only introduced when explicitly requested or when simpler parameterizations fail to achieve adequate fit. As illustrated in Figure~\ref{fig:hybrid}, neural augmentation can substantially improve fit; however, since our primary focus is on mechanistic disease modeling rather than purely predictive forecasting, we do not employ these models in the main experimental evaluation. Such neural-augmented formulations are more appropriate for forecasting-oriented tasks.

\paragraph{Selection and calibration.}
The choice among these parameterizations is guided by the scenario description and empirical performance.
Unless specified otherwise, \tool{} prioritizes time-invariant models to recover dominant epidemic trends without unnecessary complexity.
More expressive parameterizations are introduced incrementally when required to improve calibration fidelity.
Generated simulator code and corresponding fits are reported alongside projections to illustrate the effect of each design choice.

\section{Sensitivity to Language Model Capacity}
We analyze the effect of language model capacity on simulator quality (Table \ref{tab:generation_factors}, \ref{tab:model_generation_factors}, Appendix). Larger models more reliably infer epidemiological structure with weaker guidance, while smaller models depend heavily on skeleton code and explicit constraints. However, across all model sizes, the combination of structural priors, verification, and feedback is essential for producing reliable simulators.

This suggests that robustness emerges not from language model capacity alone, but from the interaction between agentic control, structural constraints, and learning.

\section{Scenario Specification and Structural Constraints for Behavioral Epidemic Baselines}

\paragraph{Scenario Specification}

Each experiment is defined by a scenario consisting of (i) location-specific epidemiological inputs and (ii) a behavioral mechanism choice. Location inputs include age-structured contact matrices, population by age, daily incident deaths (and optionally cases), Google mobility time series, hemisphere indicator for seasonality, and fixed age-specific infection fatality ratios (IFR). The behavioral mechanism specifies one of three PNAS baselines:
\begin{enumerate}
    \item DDB (Data-Driven Behavior): Exogenous modulation of transmission using mobility-derived contact reductions.
	\item CBF (Compartmental Behavior Feedback): Endogenous risk-averse behavior modeled via an additional susceptible compartment with transitions driven by observed deaths.
	\item EFB (Effective Force-of-Infection Damping): Implicit behavioral response that nonlinearly damps transmission as a function of recent and cumulative deaths.
\end{enumerate}

Scenarios do not contain equations or code; they explicitly specify assumptions and data contracts.

\paragraph{Prompt Construction and Structural Constraints}

From the scenario, EPITWIN constructs a prompt that enforces a fixed execution interface and hard scientific constraints. The prompt requires the LLM to generate a single-location, age-structured SEIR-type simulator with daily time steps, causal compartment flows $S \rightarrow E \rightarrow I \rightarrow R$, and deaths derived via fixed IFR. Transmission is defined through the force of infection
$\lambda(k,t)=s(t)\sum_{k'} C_{kk'} \frac{I_{k'}(t)}{N_{k'}} $,
with behavior affecting $\lambda(k,t)$ according to the selected baseline. To preserve epidemiological semantics, the prompt distinguishes trainable parameters (e.g., $R_0$, incubation and removal rates, detection rate, behavioral response strengths) from fixed inputs (contact matrices, populations, IFR, mobility series, seasonality phase, reporting delay). Numerical correctness (non-negativity, mass conservation, monotonic cumulative deaths) is enforced structurally rather than via ad-hoc clamping.

%% file: limitations.tex
\section{Limitations and Future Work}
\begin{enumerate}
    \item \textbf{Fixed Optimization Strategy.}
    \tool{} uses a Fixed Optimization Strategy, which prevents the agent from adaptively modifying learning rates during calibration. Future iterations will explore convergence-aware stopping criteria to prevent regression in model quality during later generations due to compounding noise.
    \item \textbf{Iterative Code Generation Stability.}
    Although iterative code refinement generally improves model quality, later generations may occasionally regress due to compounding generation noise. Similarly, automatic error recovery is bounded by a fixed retry limit to ensure termination. Extending the framework with convergence-aware stopping criteria remains an important direction for future work.
\end{enumerate}
